\pdfoutput=1

\documentclass[11pt]{article}
\usepackage{EMNLP2022}

\usepackage{times}
\usepackage{latexsym}
\usepackage{multirow}
\usepackage{graphicx}
\usepackage{booktabs}
\usepackage{amsmath}
\usepackage{xspace}
\usepackage{color}

\usepackage{enumitem}
\setitemize[1]{itemsep=0pt,partopsep=0pt,parsep=\parskip,topsep=5pt,leftmargin=10pt}

\usepackage[T1]{fontenc}

\usepackage[utf8]{inputenc}

\usepackage{microtype}

\usepackage{inconsolata}

\newcommand\ourdata{$\mathcal{C}{om}\mathcal{F}{act}$}
\newcommand\personachat{\textsc{Persona-Chat}}
\newcommand\rocatomic{\textsc{Roc-Atomic}}
\newcommand\personaatomic{\textsc{Persona-Atomic}}
\newcommand\mutualatomic{\textsc{Mutual-Atomic}}
\newcommand\movieatomic{\textsc{Movie-Atomic}}
\newcommand\atomicTT{\textsc{Atomic}$_{20}^{20}$}
\newcommand\atomic{\textsc{Atomic}}
\newcommand\comet{\textsc{Comet}}
\newcommand\eg{\textit{e.g.}}
\newcommand\ie{\textit{i.e.}}

\title{\ourdata: A Benchmark for Linking \\ Contextual Commonsense Knowledge}

\author{\textbf{Silin Gao$^{1}$, Jena D. Hwang$^{2}$, Saya Kanno$^{3}$, Hiromi Wakaki$^{3}$,} \\
\textbf{Yuki Mitsufuji$^{3}$, Antoine Bosselut$^{1\dagger}$} \\
$^1$NLP Lab, IC, EPFL, Switzerland, $^2$Allen Institute for AI, WA, USA \\
$^3$Sony Group Corporation, Tokyo, Japan \\
{\tt $^1$\{silin.gao,antoine.bosselut\}@epfl.ch, $^2$jenah@allenai.org,} \\
{\tt $^3$\{saya.kanno,hiromi.wakaki,yuhki.mitsufuji\}@sony.com}
}

\begin{document}
\maketitle
\renewcommand{\thefootnote}{\fnsymbol{footnote}}
\footnotetext[2]{Corresponding author.}
\renewcommand{\thefootnote}{\arabic{footnote}}

\begin{abstract}
Understanding rich narratives, such as dialogues and stories, often requires natural language processing systems to access relevant knowledge from commonsense knowledge graphs. 
However, these systems typically retrieve facts from KGs using simple heuristics that disregard the complex challenges of identifying situationally-relevant commonsense knowledge (\eg, contextualization, implicitness, ambiguity).

In this work, we propose the new task of commonsense fact linking, where models are given contexts and trained to identify situationally-relevant commonsense knowledge from KGs. Our novel benchmark, \ourdata{}, contains $\sim$293k in-context relevance annotations for commonsense triplets across four stylistically diverse dialogue and storytelling datasets. 
Experimental results confirm that heuristic fact linking approaches are imprecise knowledge extractors. Learned fact linking models demonstrate across-the-board performance improvements ($\sim$34.6\% F1) over these heuristics. Furthermore, improved knowledge retrieval yielded average downstream improvements of 9.8\% for a dialogue response generation task. However, fact linking models still significantly underperform humans, suggesting our benchmark is a promising testbed for research in commonsense augmentation of NLP systems.\footnote{We release our data and code to the community at \url{https://github.com/Silin159/ComFact}}
\end{abstract}

\section{Introduction}
\label{sec:intro}

\begin{figure}[t]
\centering
\includegraphics[width=1.0\columnwidth]{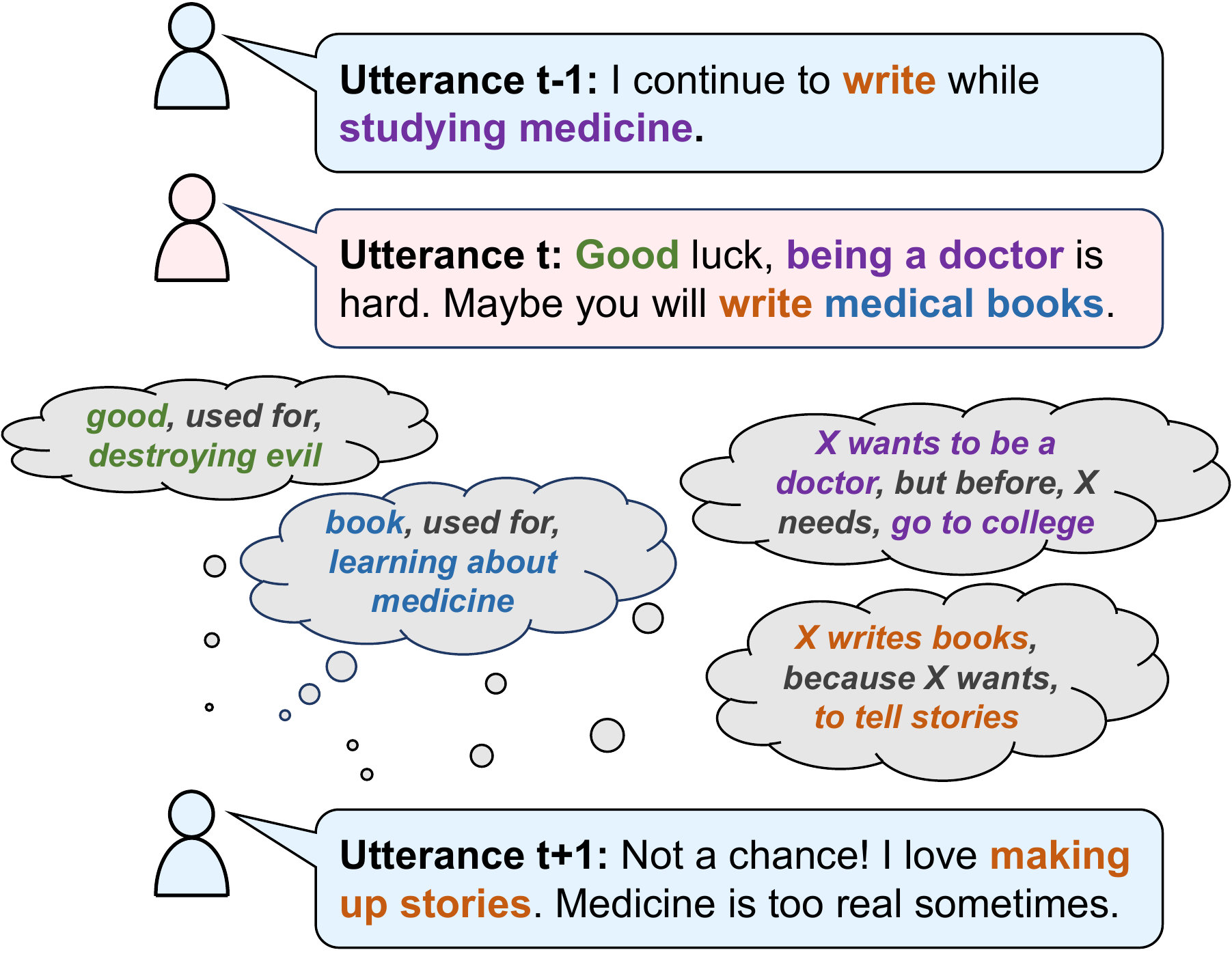}
\caption{Commonsense fact linking in a conversation.
Triples in bubbles represent linked facts.
Words and phrases in \textcolor{green!40!black}{green}, \textcolor{teal!70!blue}{blue}, \textcolor{violet!80!black}{purple} and \textcolor{orange!70!black}{orange} illustrate four different linking relationships for facts (\S\ref{ssec:data:relevance}).} 
\label{linking}
\end{figure}

In conversations, stories, and other varieties of narratives, language users systematically elide information that readers (or listeners) reliably fill in with world knowledge. For example, in Figure~\ref{linking}, the speaker of utterance $t$ (\ie, \textcolor{pink!75!gray}{pink}) infers that their counterpart (\textcolor{cyan!70!gray}{cyan}) wants to be a doctor because they are studying medicine, even though the \textcolor{cyan!70!gray}{cyan} speaker does not explicitly mention their career goals. 
To reflect this ability, language understanding systems are often augmented with knowledge bases (KBs, \eg,  \citealp{speer2017conceptnet}) that allow them to access relevant background knowledge.

Considerable research has examined how to construct large databases of world knowledge for this purpose \citep{Lenat1995CYCAL,Suchanek2007YagoAC,speer2017conceptnet,sap2019atomic}, as well as how to design models that can reason over relevant subsets of this knowledge to form a richer understanding of language (\eg, \citealp{lin2019kagnet}). However, less work examines how to retrieve these inferences (or facts) from the KB in the first place. Current methods typically rely on pattern-based heuristics \citep{mihaylov-frank-2018-knowledgeable,feng2020scalable}, unsupervised scoring using corpus statistics \citep{Weissenborn2018DynamicIO} or neural re-rankers \citep{yasunaga2021qa}, or combinations of these methods \citep{Bauer2018CommonsenseFG}. 

These simple methods produce computationally tractable knowledge representations, but frequently retrieve noisy information that is irrelevant to the narrative they are constructed to represent. Recent work demonstrates that models trained with heuristically-retrieved commonsense knowledge learn simplified reasoning patterns \citep{Wang2021GNNIA} and provide false notions of interpretability \citep{Raman2021LearningTD}. We posit that inadequate retrieval from large-scale knowledge resources is a key contributor to the spurious reasoning abilities learned by these systems. 

Acknowledging the importance of retrieving relevant commonsense knowledge to augment models, we identify a set of challenges that commonsense knowledge retrievers must address. First, retrieved commonsense knowledge must be \textbf{contextually-relevant}, rather than generically related to the entities mentioned in the context. Second, relevant commonsense knowledge can often be \textbf{implicit}, \eg, in Figure~\ref{linking}, writing may be a leisure hobby for the {\textcolor{cyan!70!gray}{cyan}} speaker, explaining why they ``love making up stories''. Finally, knowledge may be \textbf{ambiguously} relevant to a context. The {\textcolor{cyan!70!gray}{cyan}} speaker in Figure~\ref{linking} may write as a relaxing hobby, or be thinking of quitting medical school to pursue a career as a writer. Without knowing the rest of the conversation, both inferences are potentially valid.

To more adequately address these challenges, we introduce the new task of commonsense fact linking,\footnote{We follow prior naming convention for entity linking \citep{Ling2015DesignCF} and multilingual fact linking \citep{Kolluru2021MultilingualFL}, though the task can also be viewed as information retrieval (IR) from a commonsense knowledge base.} 
where models are given contexts and trained to identify situationally-relevant commonsense knowledge from KGs. 
For this task, we construct a \textbf{Com}monsense \textbf{Fact} linking dataset (\ourdata{}) to benchmark the next generation of models designed to improve commonsense fact retrieval. \ourdata{} contains $\sim$293k contextual relevance annotations for four diverse dialogue and storytelling corpora. 
Our empirical analysis shows that heuristic methods over-retrieve many unrelated facts, yielding poor performance on the benchmark. Meanwhile, models trained on our resource are much more precise extractors with an average 34.6\% absolute F1 boost (though they still fall short of human performance). The knowledge retriever developed on our resource also brings an average 9.8\% relative improvement on a downstream dialogue response generation task. These results demonstrate that \ourdata{} is a promising testbed for developing improved fact linkers that benefit downstream NLP applications.

\section{Related Work}

\paragraph{Commonsense Knowledge Graphs}
Commonsense knowledge graphs (KGs) are standard tools for providing background knowledge to models for various NLP tasks such as question answering \citep{talmor2019commonsenseqa,sap2019social} and text generation \citep{lin2020commongen}. 
ConceptNet \citep{liu2004conceptnet,speer2017conceptnet}, a commonly used commonsense KG, contains high-precision facts collected from crowdsourcing \citep{singh2002open} and web ontologies \cite{miller1995wordnet,lehmann2015dbpedia}, but is generally limited to taxonomic, lexical and physical relationships \citep{davis2015commonsense,sap2019atomic}.
\atomic{} \citep{sap2019atomic} and \textsc{Anion} \citep{Jiang2021Anion} are fully crowdsourced, and focus on representing knowledge about social interactions and events.
\atomicTT{} \citep{hwang2021comet} expands on \atomic{} by annotating additional event-centered relations and integrating the facts from ConceptNet that are not easily represented by language models, yielding a rich resource of complex entities.
In this work, we construct our \ourdata{} dataset based on the most advanced \atomicTT{} KG.

\paragraph{Commonsense Fact Linking}
Knowledge-intensive NLP tasks are often tackled using commonsense KGs to augment the input contexts provided by the dataset \citep{wang2019improving,ye2019align,gajbhiye2021exbert,yin2022survey}. Models for various NLP applications benefit from this fact linking, including question answering \citep{feng2020scalable,yasunaga2021qa,zhang2022greaselm}, dialogue modeling \citep{zhou2018commonsense,wu2020diverse} and story generation \citep{guan2019story,ji2020language}.
All above works typically conduct fact linking using heuristic solutions.

Recent research explores unsupervised learning approaches for improving on the shortcomings of heuristic commonsense fact linking. 
\citet{huang2021improving} and \citet{zhou2021think} use soft matching based on embedding similarity to link commonsense facts with implicit semantic relatedness. 
\citet{guan2020knowledge} use knowledge-enhanced pretraining to implicitly incorporate commonsense facts into narrative systems, but their approach reduces the controllability and interpretability of knowledge integration. 
Finally, several works \citep{Arabshahi2021ConverationalMH,Bosselut2019DynamicKG,peng2021inferring,peng2021guiding,tu2022misc} use knowledge models \citep{bosselut2019comet,Da2020AnalyzingCE,west2021symbolic} to generate commonsense facts instead of linking from knowledge graphs.
However, the contextual quality of generated facts from knowledge models is also under-explored in these application scenarios.
In this paper, we conduct more rigorous study on commonsense fact linking.

\section{\ourdata{} Construction}
\label{sec:comfact}
In this section, we give an overview of commonsense fact linking and its associated challenges, and describe our approach for building the \ourdata{} dataset centered around these challenges.

\subsection{Overview}
\paragraph{Notation}
We are given narrative samples $\mathcal{S}$ (\eg{}, a dialogue or story snippet) containing multiple statements (or utterances for dialogues) $[U_{1},U_{2},...,U_{T}]$.
For the $t\text{-th}$ statement $U_{t}$, the collections of statements that comprise its past and future context are defined as $U_{<t}=[U_{t-k},...,U_{t-1}]$ and $U_{>t}=[U_{t+1},...,U_{t+l}]$, respectively.

A commonsense knowledge graph $\mathcal{G}$ is made up of a set of interconnected commonsense facts, each represented as a triple containing a head entity, a tail entity, and a relation connecting them, as depicted in Figure~\ref{linking}. The task in this work is to identify the subset of commonsense facts from $\mathcal{G}$ that may be relevant for understanding the situation described in the context $C_{t}=[U_{<t},U_{t},U_{>t}]$.

\paragraph{Challenges} The task of commonsense fact linking poses several challenges:

\begin{itemize}
    \item \textbf{\textit{Contextualization}}: many facts linked using simple heuristic methods, such as string-matching, are not actually relevant to the situation described in a context. For example, in Figure~\ref{linking}, the facts in bubbles are all pattern-matched to the dialogue, but (\textit{good}, \textit{used for}, \textit{destroying evil}) turns out to not be relevant to the situation when someone says \textit{good luck}. Our study shows that only $\sim25\%$ of facts linked through string matching end up being fully relevant to the context.
    \item \textbf{\textit{Implicitness}}: some facts are linked to the context in implicit ways. For example, in Figure~\ref{linking}, the fact with \textit{go to college} is implicitly linked to the phrase \textit{studying medicine}, which makes it relevant to the context even though no direct reference to \textit{college} is made in the dialogue, precluding it from being linked using string-matching.
    \item \textbf{\textit{Ambiguity}}: different observers can disagree on on whether a fact is relevant to reason about a situation, particularly if the future context of a narrative is unknown.  
    For example, (\textit{X writes books}, \textit{because X wants}, \textit{to tell stories}) in Figure~\ref{linking} is relevant to the final produced utterance, but would not be if the final utterance had been about wanting to write scientific research papers instead (\textit{n.b.}, the best use of writing skill). 
\end{itemize}

\noindent While many methods have been proposed for linking facts in $\mathcal{G}$ to $C_{t}$, these methods typically rely on rule-based heuristics or unsupervised scoring methods, which do not adequately address the unique challenges of this task. In the following sections, we present our approach for building the \ourdata{} dataset that addresses the above challenges.

\begin{figure*}[t]
\centering
\includegraphics[width=1.0\textwidth]{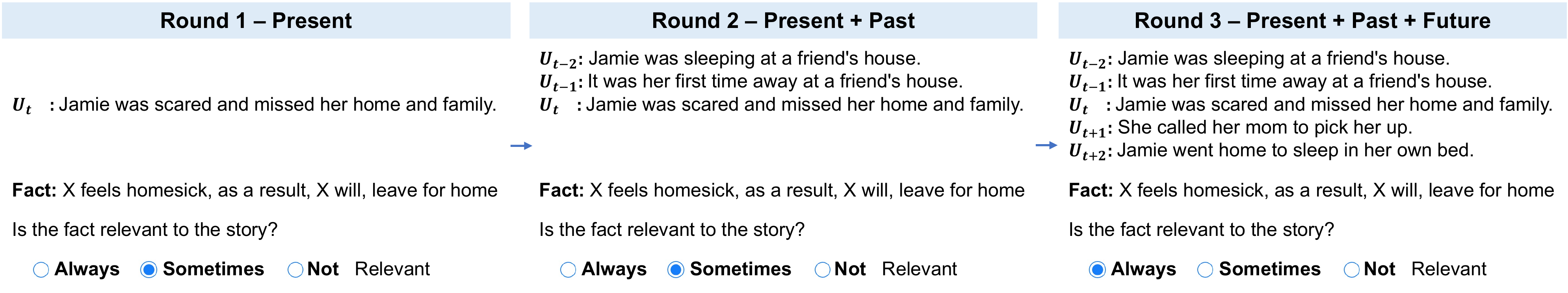}
\caption{Illustration of our three-round fact candidate validation 
}
\label{crowdsourcing}
\end{figure*}

\subsection{Fact Candidate Linking}
\label{sec:hfcl}
Given $C_{t}=[U_{<t},U_{t},U_{>t}]$ from a natural language sample $\mathcal{S}$, we link an initial set of potentially relevant fact candidates from $\mathcal{G}$ using two approaches, 
one designed to extract explicit relevant facts and one designed for implicitly relevant facts.

\paragraph{Extracting Fact Candidates}
Similar to prior works (\eg, \citealp{feng2020scalable}), we use surface-form pattern matching to retrieve head entities in $\mathcal{G}$ that are explicitly linked to $U_{t}$, and collect facts that contain the retrieved head entities as candidates.
In particular, we lemmatize and part-of-speech (POS) tag $U_{t}$ and every head entity in $\mathcal{G}$.
Then, we match patterns between these sources that are words that are informative parts of speech (\eg, nouns, verbs, adjectives, adverbs) or that correspond to $n$-grams in a master list of English idioms from Wiktionary.\footnote{\url{https://en.wiktionary.org/w/index.php?title=Category:English_idioms}} 
We retrieve head entities whose informative patterns all appear in the set of patterns from $U_{t}$.

However, pattern matching only extracts a set of fact candidates whose head entities can be explicitly recovered from the context $U_t$. To retrieve facts that may be semantically related to the context, but cannot be explicitly linked through patterns (\eg{}, paraphrased facts), we use embedding similarity matching \citep{zhou2021think}.
In particular, we use Sentence-BERT \citep{reimers2019sentence} to encode $U_{t}$ along with every head entity in $\mathcal{G}$ as embedding vectors, and select the $\text{top-}5$ head entities whose embeddings have the highest cosine similarity with the embedding of $U_{t}$. Using this approach, we extend the sets of available candidates often retrieved by pattern matching methods and include implicit inferences in our candidate set.

\paragraph{Filtering Fact Candidates}
Head entities linked via pattern and embedding matching may connect to tail entities whose semantics are far different from that of $C_{t}$ (\eg{}, \textit{destroying evil} in Figure~\ref{linking}).
Consequently, we perform a first round of automatic filtering by pruning the tail entities of each head entity according to their similarity to $C_{t}$. Using Sentence-BERT, we encode each tail entity and $C_{t}$ as embedding vectors.
For each head entity, we keep its $\text{top-}5$ tail entities that have the highest embedding cosine similarity with that of $C_{t}$.

\subsection{Crowdsourcing Relevance Judgements}
\label{sec:cvv}
We use the prior heuristics to over-sample a large initial set of knowledge ($\sim$46 facts per example context). We then devise a two-step procedure for evaluating the contextual relevance of these linked fact candidates using crowdworkers from Amazon Mechanical Turk, which we describe below.

\begin{figure}[t]
\centering
\includegraphics[width=1.0\columnwidth]{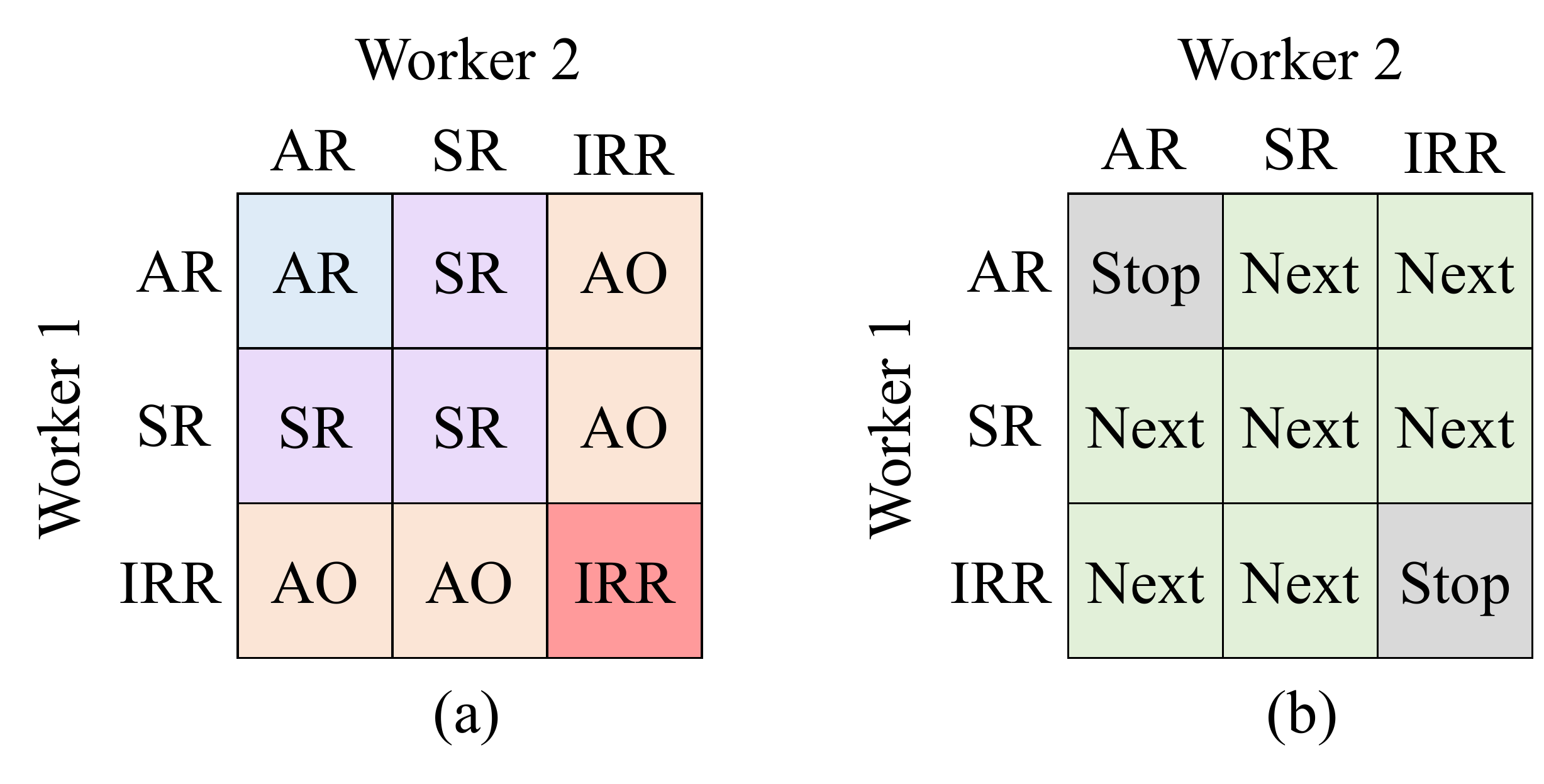}
\caption{Summary of rules in fact candidate validation rounds. (a) Mapping from worker annotations to relevance labels: \textit{always relevant} (AR), \textit{sometimes relevant} (SR), \textit{at odds} (AO) and \textit{irrelevant} (IRR). (b) Mapping from worker annotations to action of the round:  \textit{evaluate in the next round} (Next) and \textit{end validation} (Stop).}
\label{mapping}
\end{figure}

\paragraph{Validating Head Entities}
First, we task workers with validating the relevance of head entities with respect to the context.
For each head entity, we show two workers $C_{t}$ and a head candidate associated with $U_{t}$, and independently ask them to judge whether the head candidate is relevant to $U_{t}$.
Head candidates are labeled as: a) \textbf{\textit{relevant} with full confidence} if both workers identify the head entity as relevant, b) \textbf{\textit{relevant} with half confidence} if only one of the workers choose relevant, or c) \textbf{\textit{irrelevant}} if neither of the workers choose relevant.

\begin{table*}[t]
\centering
\resizebox{1.0\textwidth}{!}{
\smallskip\begin{tabular}{@{~}l@{~~~}c@{~~~}c@{~~~}c@{~~~}c@{~~~}c@{~~~}c@{~~~}c@{~~~}c@{~~~}c@{~~~}c@{~~~}c@{~~~}c@{~~~}c@{~~~}c@{~~~}c@{~~~}c@{~~~}c@{~~~}c@{~~~}c@{~~~}c@{~}}
\toprule
\multirow{2}*{\textbf{Method}} & \multicolumn{5}{c}{\personaatomic{}} & \multicolumn{5}{c}{\mutualatomic{}} & \multicolumn{5}{c}{\rocatomic{}} & \multicolumn{5}{c}{\movieatomic{}}\\
                        \cmidrule(lr){2-6}  \cmidrule(lr){7-11} \cmidrule(lr){12-16} \cmidrule(lr){17-21}
                        & \textbf{AR} & \textbf{SR} & \textbf{AO} & \textbf{IRR} & {$\kappa$} & \textbf{AR} & \textbf{SR} & \textbf{AO} & \textbf{IRR} & {$\kappa$} & \textbf{AR} & \textbf{SR} & \textbf{AO} & \textbf{IRR} & {$\kappa$} & \textbf{AR} & \textbf{SR} & \textbf{AO} & \textbf{IRR} & {$\kappa$} \\
\toprule
\multirow{2}*{Explicit} &  8042   &  772  &  3731  & 20940  & \multirow{2}*{0.72} &  5352  &  910  &  4007  &  6541  & \multirow{2}*{0.52}  &  8981  &  862  &  4449  &                         20320  & \multirow{2}*{0.69} &  7352  &  1883  &  9578  &  23957  & \multirow{2}*{0.56} \\
                        &  24\%   &  2\%  &  11\%  &  63\%  &                     &  32\%  &  5\%  &  24\%  &  39\%  &                      &  26\%  &  2\%  &  13\%  &  59\%  &                     &  17\%  &  5\%  &  22\%  &  56\%  &                      \\
\midrule
\multirow{2}*{Implicit} &  2277  &  224  &  1076  & 3813   & \multirow{2}*{0.68}  &  4206  &  736  &  2921  &  4003  & \multirow{2}*{0.49}  &  6068  &  653  &  3177  &                          9234  & \multirow{2}*{0.64} &  2582  &  635  &  2862  &  5717  & \multirow{2}*{0.55} \\
                        &  31\%  &  3\%  &  15\%  &  51\%  &                      &  35\%  &  6\%  &  25\%  &  34\%  &                      &  32\%  &  3\%  &  17\%  &  48\%  &                     &  22\%  &  5\%  &  24\%  &  49\%  &                      \\
\midrule
\multirow{2}*{Both}     & 10319  &  996  &  4807  & 24753  & \multirow{2}*{0.71}  &  9558  &  1646  &  6928  &  10544  & \multirow{2}*{0.51}   &  15049 & 1515  &  7626  &                         29554  & \multirow{2}*{0.67} &  9934  &  2518  &  12440 &  29674  & \multirow{2}*{0.56} \\
                        &  25\%  &  2\%  &  12\%  &  61\%  &                      &  33\%  &   6\%  &  24\%  &  37\%  &                       &  28\%  &  3\%  &  14\%  &  55\%  &                     &  18\%  &   5\%  &  23\%  &  54\%  &                      \\
\bottomrule
\end{tabular}
}
\caption{Relevance of fact candidates for different candidate extraction methods. \textbf{AR}: \textit{always relevant}, \textbf{SR}: \textit{sometimes relevant}, \textbf{AO}: \textit{at odds}, \textbf{IRR}: \textit{irrelevant}. {$\kappa$} denotes the Cohen's $\kappa$.}
\label{tab:fact_stats}
\end{table*}

\paragraph{Validating Fact Candidates}
After curating a set of relevant head entities, workers then validate the relevance of the fact candidates associated with those head entities.\footnote{If a head entity is deemed \textbf{\textit{irrelevant}}, we assume that all fact candidates associated with it are \textit{irrelevant} as well.}
To evaluate contextual relevance of facts in a fine-grained manner, we define a three-round task for workers, as shown in Figure~\ref{crowdsourcing}.

In the first round, we show two workers $U_{t}$ and the set of fact candidates, and independently ask them to judge whether the fact candidate is \textit{always relevant}, \textit{sometimes relevant}, or \textit{irrelevant} to $U_t$.\footnote{From feedback, we observe that crowdworkers prefer our fine-grained annotation scheme as it allows them to express uncertainty in the judgment compared to a binary choice.}
In the second round, we repeat this task, but show the past context along with $U_t$, namely $[U_{<t},U_{t}]$.
In the third round, we repeat the task again, but show the full context $C_{t}=[U_{<t},U_{t},U_{>t}]$.

After each round, we assign or update the relevance label of a fact candidate as: a) \textbf{\textit{always relevant}} if both workers label it \textit{always relevant}, b) \textbf{\textit{sometimes relevant}} if one or both of the workers label it \textit{sometimes} instead of \textit{always} \textit{relevant}, c) \textbf{\textit{at odds}} if one worker chooses \textit{always} or \textit{sometimes} \textit{relevant} and the other chooses \textit{not relevant}, d) \textbf{\textit{irrelevant}} if both workers select \textit{not relevant} (as shown in Figure~\ref{mapping}a).
In practice, we find that including more context (\ie{}, $U_{<t}$ or $U_{>t}$) rarely changes the validation of an initially \textit{always relevant} or \textit{irrelevant} fact. 
So after each round, if a fact candidate is labeled as \textit{always relevant} or \textit{irrelevant}, we do not evaluate it in the next round. Otherwise, there is relevance ambiguity over a fact, and we validate it again in the next round with additional context (as shown in Figure~\ref{mapping}b). In the second and third rounds, if a worker annotates a fact candidate as \textit{always} or \textit{sometimes} \textit{relevant}, we ask them to justify their selection by identifying which statement(s) in $U_{<t}$ or $U_{>t}$ make the fact relevant to the situation. 

\subsection{Fine-grained Contextual Relevance} 
\label{ssec:data:relevance}
The three rounds of assessment allow us to perform a fine-grained annotation of the contextual relevance of a fact candidate.
For each fact candidate, we map its relatedness to $C_{t}=[U_{<t},U_{t},U_{>t}]$ to one of the following four link types:
\begin{itemize}
    \item \textbf{R}elevant to \textbf{P}resent \textbf{A}lone (\textbf{RPA}): the linked fact is directly relevant to $U_{t}$ alone.
    For example, in Figure~\ref{linking}, the fact highlighted in \textcolor{teal!70!blue}{blue}, \textit{medical books} are used for \textit{learning about medicine}, is relevant to the concept of medical books mentioned in the utterance.
    
    \item \textbf{R}elevant to \textbf{P}resent given the \textbf{P}ast (\textbf{RPP}): the linked fact is not relevant to $U_{t}$ alone, but relevant to $U_{t}$ given $U_{<t}$.
    As shown in Figure~\ref{linking}, the \textcolor{violet!80!black}{purple} fact helps interpret that \textit{studying medicine} happens when someone \textit{goes to college}, which is also a prerequisite of \textit{being a doctor}.
    
    \item \textbf{R}elevant to \textbf{P}resent given the \textbf{F}uture (\textbf{RPF}): the linked fact is to $U_{t}$ knowing $U_{<t}$ and $U_{>t}$.
    For example, the fact colored in \textcolor{orange!70!black}{orange} in Figure~\ref{linking} helps associate the action \textit{Person X writes books} with the reason \textit{making up stories}.
    
    \item \textbf{IRR}elevant to $C_{t}$ (\textbf{IRR}): the linked fact is not relevant to the situation described in $C_{t}$.
    For example, as shown in Figure~\ref{linking}, the fact colored in \textcolor{green!40!black}{green} is irrelevant to help understand the context, although it is linked to \textit{good luck} in surface form.
\end{itemize}
If a fact candidate is finally labeled as \textit{irrelevant}, we label its link type as \textbf{IRR}.
If a fact candidate is finally labeled as \textit{at odds}, we do not label its link type since its relevance is controversial.
Otherwise, we further check the earliest assessment round where the fact's final relevance label comes out: we label the fact's link type as \textbf{RPA}, \textbf{RPP}, or \textbf{RPF} if its final relevance label first comes out at the first, second or third round, respectively.

\paragraph{Retaining Disagreements} 
Facing the challenge of ambiguity in relevance (potentially due to inherent uncertainty in the facts being linked; \citealp{pavlick-kwiatkowski-2019-inherent}), we track disagreements between workers throughout our annotation pipeline, allowing us to measure the relevance controversy in commonsense fact linking.
In particular, we record the disagreements of workers when: a) a head entity is \textbf{\textit{relevant with half confidence}}, b) a fact candidate is \textbf{\textit{sometimes relevant}}, c) a fact candidate is \textbf{\textit{at odds}} in relevance.\footnote{For a fact candidate, we record judgements from the round where the fact's final relevance label first comes out.} These rich annotations enable multiple modeling settings at different granularities for identifying relevant inferences, providing a rich set of potential label spaces for future work in granular fact linking.


\begin{table*}[t]
\centering
\resizebox{1.0\textwidth}{!}{
\smallskip\begin{tabular}{lcccccccccccccccc}
\toprule
\multirow{2}*{\textbf{Relevance}} & \multicolumn{4}{c}{\personaatomic{}} & \multicolumn{4}{c}{\mutualatomic{}} & \multicolumn{4}{c}{\rocatomic{}} & \multicolumn{4}{c}{\movieatomic{}} \\
                           \cmidrule(lr){2-5}         \cmidrule(lr){6-9}       \cmidrule(lr){10-13}       \cmidrule(lr){14-17}
 & \textbf{RPA}  & \textbf{RPP} & \textbf{RPF} & \textbf{all} & \textbf{RPA} & \textbf{RPP} & \textbf{RPF} & \textbf{all} & \textbf{RPA} & \textbf{RPP} & \textbf{RPF} & \textbf{all} & \textbf{RPA} & \textbf{RPP} & \textbf{RPF} & \textbf{all} \\
\toprule
\multirow{2}*{Always}    &  8310  &  1272  &  738  & 10320 &  6678  &  1728  &  1152  & 9558  & 12048 &  1562   &  1439  & 15049 &  6495  &  1681  &  1758  &  9934 \\
                         &  81\%  &  12\%  &  7\%  & 100\% &  70\%  &  18\%  &  12\%  & 100\% &  80\% &   10\%  &  10\%  & 100\% &  65\%  &  17\%  &  18\%  & 100\% \\
\midrule
\multirow{2}*{Sometimes} &  523   &  130   &  342  &  995  &  801   &  316   &  529   & 1646  &  734  &   275   &  506   & 1515  &  1132  &  262   &  1124  &  2518 \\
                         &  53\%  &  13\%  &  34\% & 100\% &  49\%  &  19\%  &  32\%  & 100\% &  48\% &   18\%  &  34\%  & 100\% &  45\%  &  10\%  &  45\%  & 100\% \\
\midrule
\multirow{2}*{Both}      &  8833  &  1402  &  1080 & 11315 &  7479  &  2044  &  1681  & 11204 & 12782 &  1837   &  1945  & 16564 &  7627  &  1943  &  2882  & 12452 \\
                         &  78\%  &  12\%  &  10\% & 100\% &  67\%  &  18\%  &  15\%  & 100\% &  77\% &   11\%  &  12\%  & 100\% &  61\%  &  16\%  &  23\%  & 100\% \\
\bottomrule
\end{tabular}
}
\caption{Link type statistics of relevant facts on each data subset of \ourdata{}.}
\label{tab:link_stats}
\end{table*}

\begin{table}[t]
\centering
\resizebox{1.0\columnwidth}{!}{
\smallskip\begin{tabular}{l}
\hline
\multicolumn{1}{c}{\textbf{Context}}\\
\hline
$U_{t-2}$: I like cooking macrobiotic and healthy food\\
\quad\quad\; and working out at the gym.\\
$U_{t-1}$: What is macrobiotic food? My best friend is my mother.\\
$U_{t}$\quad: Things like whole grains. I drink at bars,\\
\quad\quad\; so I have to stay healthy.\\
$U_{t+1}$: You should not drink a lot, it's bad for you.\\
$U_{t+2}$: Well that is where I meet women, at bars.\\
\quad\quad\; So I end up drinking.\\
\hline
\multicolumn{1}{c}{\textbf{Facts}}\\
\hline
\textbf{RPA}: stay healthy, HasSubEvent, eat healthy foods\\
\quad\quad\; (\textit{always relevant})\\
\textbf{RPP}: stay healthy, xNeed, exercise and eat balanced meals\\
\quad\quad\; (\textit{always relevant})\\
\textbf{RPF}: bar, ObjectUse, take their friends to\\
\quad\quad\; (\textit{sometimes relevant})\\
\textbf{IRR}\,: PersonX likes to drink, xAttr, thirsty\\
\quad\quad\; (\textit{irrelevant})\\
\hline
\end{tabular}
}
\caption{\personaatomic{} example annotations.}
\label{tab:example_persona}
\end{table}

\section{\ourdata{} Analysis}
\label{data_analysis}
We use \atomicTT{} \citep{hwang2021comet} as the commonsense KG for building \ourdata{}, which contains 1.33M complex facts covering physical objects, daily events and social interactions.  \atomicTT{} is a rich resource for building our dataset, as it covers rich knowledge types (\eg, physical, social, and event knowledge), and is partially consolidated from other popular KGs including ConceptNet \citep{speer2017conceptnet}) and \atomic{} \citep{sap2019atomic}, potentially offering better generalization for fact linking with these other resources.

We sample narrative contexts from four stylistically diverse English dialogue and storytelling datasets that involve elaborate contextual inference and understanding: \personachat{} \cite{zhang2018personalizing}, MuTual \citep{cui2020mutual}, ROCStories \cite{mostafazadeh2016corpus} and the CMU Movie Summary Corpus \citep{bamman2013learning}.
The context window size is set to 5 where $U_{<t}=[U_{t-2},U_{t-1}]$ and $U_{>t}=[U_{t+1},U_{t+2}]$.
We denote the data portions collected from the four datasets as \personaatomic{}, \mutualatomic{}, \rocatomic{} and \movieatomic{} in \ourdata{}.\footnote{See Appendix~\ref{apdx:collection} for more data collection details.}

\paragraph{Contextual Relevance}
Table~\ref{tab:fact_stats} shows stratified statistics of the crowdsourced fact relevance annotations for the different candidate linking methods described in Sec.~\ref{sec:hfcl} (\ie, explicit pattern matching, implicit embedding matching).
We observe that unsupervised fact linking methods, whether based on heuristics for explicit patterns or implicit matching mechanisms, often link irrelevant and unrelated facts, introducing noise to any resulting extracted knowledge representation.
Interestingly, once irrelevant head entities were removed, implicit fact candidates retrieved using embedding similarity were more likely to be judged relevant by human annotators, compared to pattern-matched fact candidates, showing the importance of generating a rich set of potentially relevant fact candidates.


To quantitatively measure the ambiguity of linked commonsense facts' relevance, we use Cohen's $\kappa$ \cite{cohen1960coefficient} to measure the agreement between workers that annotate the same facts. Most $\kappa$ scores fall within the ranges that Cohen described as ``moderate'' (0.4 - 0.6) or ``substantial'' (0.6 - 0.8) agreement. 
We do observe that implicitly linked fact candidates have lower $\kappa$ scores in their relevance validation, likely because they are linked to the context in a less straightforward way, which may lead to more subjective relevance judgements.

\begin{table*}[t]
\centering
\resizebox{0.95\textwidth}{!}{
\smallskip\begin{tabular}{cllcccccccc}
\toprule
\multirow{2}*{\textbf{Context}} & \multirow{2}*{\textbf{Model}} & \multirow{2}*{\textbf{Setting}} & \multicolumn{2}{c}{\personaatomic{}} & \multicolumn{2}{c}{\mutualatomic{}} & \multicolumn{2}{c}{\rocatomic{}} & \multicolumn{2}{c}{\movieatomic{}}\\
                                                \cmidrule(lr){4-5}   \cmidrule(lr){6-7}   \cmidrule(lr){8-9}   \cmidrule(lr){10-11}
                                 &      &       & \textbf{Acc.} & \textbf{F1} & \textbf{Acc.} & \textbf{F1} & \textbf{Acc.} & \textbf{F1} & \textbf{Acc.} & \textbf{F1} \\
\toprule
\multirow{10}*{$U_{\leq t}$} & Heuristic & \multirow{2}*{none} & 0.211 & 0.348    & 0.290 & 0.450    & 0.231 & 0.375   & 0.205 & 0.340 \\
& Head Linking                   &              & 0.600 & 0.487      & 0.643 & 0.602      & 0.537 & 0.484      & 0.548 & 0.452 \\
\cmidrule(lr){2-11}
& \multirow{1}*{LSTM}            & \multirow{5}*{direct} & 0.805 & 0.471      & 0.749 & 0.573      & 0.761 & 0.457      & 0.769 & 0.417 \\
& \multirow{1}*{DistilBERT}      &              & 0.840 & 0.626      & 0.792 & 0.649      & 0.811 & 0.608     & 0.800 & 0.512 \\
& \multirow{1}*{BERT (base)}     &              & 0.859 & 0.659      & 0.801 & 0.668      & 0.841 & 0.669     & 0.818 & 0.547 \\
& \multirow{1}*{BERT (large)}    &              & 0.859 & 0.660      & 0.819 & 0.689      & 0.848 & 0.690     & 0.835 & 0.595 \\
& \multirow{1}*{RoBERTa (base)}  &              & 0.866 & 0.674      & 0.810 & 0.682      & 0.845 & 0.687     & 0.820 & 0.553 \\
\cmidrule(lr){2-11}
& \multirow{2}*{RoBERTa (large)} & direct       & 0.883 & 0.716      & 0.835 & 0.724      & 0.874 & 0.740     & 0.850 & 0.631 \\
&                                & pipeline     & 0.874 & 0.698      & 0.819 & 0.717      & 0.861 & 0.721     & 0.834 & 0.606 \\
\cmidrule(lr){2-11}
& \multirow{1}*{DeBERTa (large)} & direct       & 0.885 & 0.717      & 0.861 & 0.766      & 0.884 & 0.763     & 0.850 & 0.651 \\
\cmidrule(lr){1-11}
\multirow{3}*{$C_{t}$} & \multirow{2}*{RoBERTa (large)} & direct   & 0.882 & 0.721   & 0.838 & 0.740   & 0.879 & 0.748   & 0.851 & 0.635 \\
                       &                                & pipeline & 0.874 & 0.693   & 0.825 & 0.722   & 0.867 & 0.731   & 0.830 & 0.603 \\
\cmidrule(lr){2-11}
& Human                          & none      & \textbf{0.936} & \textbf{0.921} & \textbf{0.934} & \textbf{0.941} & \textbf{0.962} & \textbf{0.952} & \textbf{0.933} & \textbf{0.902} \\
\bottomrule
\end{tabular}
}
\caption{Fact linking results on the four data subsets of \ourdata{}. We observe a substantial performance improvement by model-based fact linkers over heuristics typically used for fact linking. A large gap remains between the performance of best model-based fact linker (based on \textbf{DeBERTa}) and \textbf{Human} performance.}
\label{tab:main}
\end{table*}

\paragraph{Link Types}
Table~\ref{tab:link_stats} shows statistics of the fine-grained link types (\S\ref{ssec:data:relevance}) for \textit{always} and \textit{sometimes relevant} facts.
We find that \textit{always relevant} facts are mostly linked to the present statements alone (\ie{}, \textbf{RPA}), while \textit{sometimes relevant} facts are more often recognized with respect to larger context windows where the past and future statements are given (\ie{}, \textbf{RPP}, \textbf{RPF}).
Even though making up a relatively small total number, \textit{sometimes relevant} facts may be critical inferences for imagining the future of the narrative, as they provide ambiguous hypotheses for where a narrative may be heading.
In general, facts linked to the present alone occupy the largest proportion of relevant facts.\footnote{See Appendix~\ref{apdx:statistics} for more data statistic results.}

Table~\ref{tab:example_persona} shows an example dialogue context from the \personaatomic{} data portion of \ourdata{}, combined with four commonsense facts from \atomicTT{} which are linked to the context with the four link types.\footnote{See Appendix~\ref{apdx:example} for more data examples in \ourdata{}.} As described above, the \textbf{RPP} and \textbf{RPF} linked facts require different portions of the context to be known for their relevance to the original statement $U_t$ to become clear. 

\paragraph{Complex Structures} Even though our dataset contains annotations for \textit{individual} linked facts, we find that complex graphical structure emerges among these annotated facts. Each narrative sample we annotate results in an average of 101 \textit{bridge} paths where two relevant facts share the same tail entity. Such paths form potentially explanatory multi-hop reasoning chains among facts relevant to the narrative sample. We also find an average of 492 bridge paths among irrelevant facts and 143 bridge paths between relevant and irrelevant facts (\ie, likely invalid reasoning chains), demonstrating the importance of precisely retrieving facts to avoid spurious explanations \citep{Raman2021LearningTD}.
\section{Experimental Methods}
We evaluate our new benchmark using various baseline classification methods based on neural language models. 
All LMs are individually trained and evaluated on the four \ourdata{} datasets.

\paragraph{Approach}
Our models encode the concatenation of a narrative context with each of its fact candidates 
selected in Sec~\ref{sec:hfcl}.
The output hidden states of the language models are then input to a binary classifier, which predicts whether the fact candidate is relevant to the context.
We consider two context windows $U_{\leq t}=[U_{<t},U_{t}]$ and $C_{t}=[U_{<t},U_{t},U_{>t}]$ in our experiments.

\paragraph{Models}
We use various pretrained language models as encoders for classifying facts, particularly the \textbf{BERT} \citep{devlin2019bert} and \textbf{RoBERTa} \citep{liu2019roberta} model families.
We test \textit{base} and \textit{large} sizes of these models, as well as more light-weight \textbf{DistilBERT} \citep{sanh2019distilbert} and more advanced \textbf{DeBERTa} \citep{he2020deberta}. 
We also test the performance of a two-layer bi-directional \textbf{LSTM}. We evaluate each model on: a) \textbf{direct} prediction, where the model is trained to directly classify the fact candidates, and b) \textbf{pipeline} prediction, where a first model is trained to classify head entities, and a second model classifies fact candidates associated with relevant head entities.\footnote{See Appendix~\ref{apdx:exp} for more details of experimental settings.}

As a baseline, we report the performance of the same \textbf{Heuristic} used to generate candidates, which predicts all fact candidates retrieved in Sec~\ref{sec:hfcl} as relevant (\ie, the typical linking approach for many methods).
We also include a semi-heuristic baseline \textbf{Head Linking}, which finetunes RoBERTa (large) to only classify the head entities linked in Sec~\ref{sec:hfcl}, and predicts the relevance of each fact candidate as the relevance of its head. 
Finally, we run a \textbf{Human} study on a randomly sampled set of 200 contexts from each of the 4 test sets of \ourdata{}, and ask crowdworkers to judge the relevance of 3 linked facts with respect to each context.

\paragraph{Data Preprocessing}
For all head entities labeled as \textit{relevant} in Sec~\ref{sec:cvv},  (regardless of the confidence), we combine fact candidates labeled as \textit{always} and \textit{sometimes relevant} as positive samples, and keep fact candidates labeled as \textit{irrelevant} as negative samples. Fact candidates labeled as \textit{at odds} are not included in this evaluation, 
though we release them as part of the dataset. For head entities initially labeled as \textit{irrelevant}, all their fact candidates are \textit{irrelevant} too, \ie{}, negative samples.


\section{Experimental Results}
\label{results}

We report the \textbf{Accuracy} and \textbf{F1} of fact candidate classification on \ourdata{} in Table~\ref{tab:main}.\footnote{Other unpresented results yield similar conclusions. The full evaluation results are included in Appendix~\ref{apdx:full}.} 
We find that trained fact linkers significantly outperform the \textbf{Heuristic} baselines, showing supervised neural classification on top of heuristically selected facts can significantly improve fact linking quality. Furthermore, the improvement over the \textbf{Head Linking} baseline demonstrates the importance of linking facts individually, rather than relying on coarse-grained entity linking.
However, model-based fact linkers are still far from \textbf{Human} performance on \ourdata{}, demonstrating that there is still considerable room for improvement. 
Interestingly, we find that directly predicting fact links outperforms a pipeline approach that first predicts relevant head entities and then only classifies fact candidates of relevant head entities
.\footnote{See Appendix~\ref{apdx:pipe} for full evaluation results of fact linking sub-tasks in pipeline setting.}

\begin{table}[t]
\centering
\resizebox{0.9\columnwidth}{!}{
\smallskip\begin{tabular}{lrrrr}
\toprule
\textbf{Model}                & \textbf{Acc.} & \textbf{Prec.} & \textbf{Recall} & \textbf{F1} \\
\toprule
Heuristic (explicit) & 0.711 & 0.712 & 0.801 & 0.754 \\
Heuristic (implicit) & 0.289 & 0.291 & 0.199 & 0.236 \\
Heuristic            & 0.553 & 0.553 & \textbf{1.000$^*$} & 0.712 \\
\midrule
RoBERTa (large)      & \textbf{0.834} & \textbf{0.834} & 0.874 & \textbf{0.854} \\
\bottomrule
\end{tabular}
}
\caption{Head entity linking results on \personaatomic{} under the context window $U_{\leq t}=[U_{<t},U_{t}]$. $^*$Recall should be perfect here because the fact candidates for which we crowdsource relevance annotations are drawn from this heuristic.}
\label{tab:head}
\end{table}

\begin{table}[t]
\centering
\resizebox{0.85\columnwidth}{!}{
\smallskip\begin{tabular}{crrrr}
\toprule
\textbf{Context} &  \textbf{RPA}  &  \textbf{RPP}  &  \textbf{RPF}  &  \textbf{all}  \\
\toprule
$U_{\leq t}$ & 0.735 & 0.634 & 0.561 & 0.698 \\
$C_{t}$      & 0.749 & 0.653 & 0.651 & 0.720 \\
\bottomrule
\end{tabular}
}
\caption{\textbf{Recall} of relevant facts by \textbf{RoBERTa (large)} on \personaatomic{} with respect to different link types and context windows, under the direct setting.} 
\label{tab:relation}
\end{table}

\begin{figure}[t]
\centering
\includegraphics[width=1.0\columnwidth]{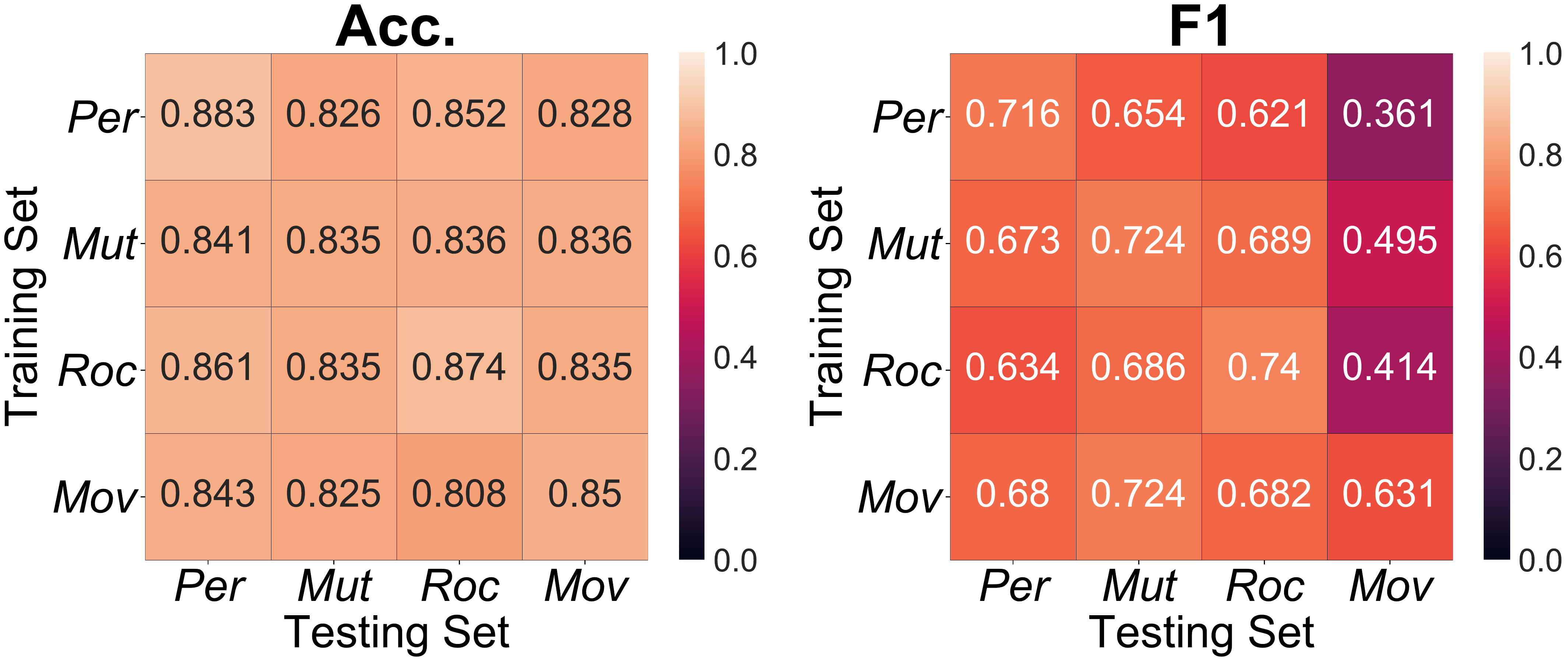}
\caption{Fact linking results of \textbf{RoBERTa (large)} across the four data portions of \ourdata{}: \personaatomic{} (\textit{Per}), \mutualatomic{} (\textit{Mut}), \rocatomic{} (\textit{Roc}) and \movieatomic{} (\textit{Mov}), under the direct setting and context window $U_{\leq t}=[U_{<t},U_{t}]$.}
\label{cross}
\end{figure}

\paragraph{Entity Linking}
Despite the error propagation, we see in Table~\ref{tab:head} that a commonsense head entity linker trained on the \personaatomic{} dataset considerably outperforms \textbf{Heuristic} baselines for entity linking too. While the \textbf{Heuristic} model can be viewed as a \textbf{Recall} oracle\footnote{
Measuring recall in knowledge retrieval is a recurring challenge as it requires a gold set of relevant facts. In our case, we record recall with respect to a gold candidate set that our heuristics initially over-sample. We provide more discussion on this decision in Appendix~\ref{apdx:exp}.} 
in our setup, the precision score of this baseline ends up being considerably worse. 
%
%
For the purpose of this analysis, we also decompose the \textbf{Heuristic} baseline into its explicit and implicit components to investigate their individual entity linking performance. 
This decomposition corresponds to the two entity retrieval methods based on explicit pattern and implicit embedding described in Section~\ref{sec:hfcl}.
Due to the link type imbalance (\ie, more explicit entity links), we find that Heuristic (explicit) recalls more of the head entities of the corpus, but still suffers in terms of precision, as many of these entities are irrelevant to the context.
Heuristic (implicit) has low precision and recall, reinforcing the challenge of identifying relevant facts that are implicit. 

\begin{table*}[t]
\centering
\resizebox{1.0\textwidth}{!}{
\smallskip\begin{tabular}{@{~}l@{~~~}cccc@{~~}c@{~~}c@{~~}c@{~}}
\toprule
\textbf{Model} & \textbf{PPL} & \textbf{Distinct-1/2} & \textbf{BLEU-1/2/3/4} & \textbf{METEOR} & \textbf{ROUGE-L} & \textbf{CIDEr} & \textbf{SkipThoughts} \\
\toprule
\textbf{CEM}            & \textbf{36.11} & \textbf{0.661}/2.998 &                  0.133/0.063/0.034/0.021                    &      0.071     &               0.163       &      0.160     &       0.478     \\
\textbf{CEM} w/ \ourdata{} &      36.14     & 0.655/\textbf{3.009} & \textbf{0.151}/\textbf{0.072}/\textbf{0.040}/\textbf{0.026} & \textbf{0.074} & \textbf{0.171} & \textbf{0.186} &  \textbf{0.496} \\
\bottomrule
\end{tabular}
}
\caption{Downstream dialogue response generation results on the EmpatheticDialogues dataset.}
\label{tab:downstream}
\end{table*}

\paragraph{Contextual Prediction}
Commonsense fact linking may be used to identify generic inferences in KGs that help augment full contexts, or to provide inferences that could help generate future portions of a narrative. To simulate fact linking for these two settings, we run experiments for two different input contexts: $C_{t}$, where the full context is given to the fact linker, and $U_{\leq t}$, where only the present and past context is given.

As expected, our results show that models that receive the half-context window $U_{\leq t}$ perform worse than those that receive the full context window $C_{t}$ as input, largely due to not recovering facts associated with the future context. In Table~\ref{tab:relation}, we observe a decreasing recall from \textbf{RPA} to \textbf{RPP} to \textbf{RPF} facts, showing that linking is more difficult when a commonsense fact requires a more challenging contextual relevance judgment. The performance degradation supports our contention on the temporal \textit{ambiguity} of commonsense facts: without knowing the future, many possible inferences may seem relevant or irrelevant in the present. 
However, despite their importance, these facts that are relevant to the future (\textbf{RPF}) make up a small portion of the data (Table~\ref{tab:link_stats}), so this shortcoming is not as clear in the overall reported performance.

\paragraph{Cross-Resource Generalization}
As annotating new commonsense fact links for all narrative datasets would be too expensive, fact linkers will need to generalize to new types of contexts with minimal performance loss. Consequently, we evaluate the performance of our fact linkers across different training and testing combinations of the four \ourdata{} data portions, with results for RoBERTa (large) shown in Figure~\ref{cross}. 

Optimistically, we find that the model trained on \movieatomic{} generalizes reasonably well to the other data portions. Models trained on the other data subsets still significantly beat the heuristic baselines on all testing subsets (and many of the other baselines from Table~\ref{tab:main}), but do not transfer as robustly. In particular, \movieatomic{} poses a challenging adaption problem, likely due to the relatively longer and more complex narratives found in the MovieSummaries corpus on which \movieatomic{} is annotated.\footnote{See Appendix~\ref{apdx:example} for data examples in \ourdata{}.} 
Therefore, our trained fact linkers still have room to improve before they reliably scale to open-domain narrative corpora, making \ourdata{} a promising testbed for further research on developing scalable fact linkers.

However, narrative generalization may not be enough for scaling fact linkers. As other knowledge resources are available (and new ones are constructed), models trained on our data should generalize to link to new commonsense knowledge resources. While more research is needed into cross-KG fact linking, we note that since \ourdata{} is developed with \atomicTT{}, models trained on our benchmark learn to link rich physical, event-based, and social interaction commonsense inferences. In fact, as a portion of \atomicTT{} includes a subset of ConceptNet \citep{speer2017conceptnet}, we can stratify performance along the inferences found in ConceptNet (mainly physical relations), and see that our models actually perform better on this subset of the data than on social interactions from \atomic.\footnote{See Appendix~\ref{apdx:knowledge_type} for more analysis on knowledge graph generalization of fact linkers.}

\paragraph{Downstream Application} Our resource enables the development of improved fact linking models, which will provide more contextually-relevant commonsense knowledge to downstream task systems. To evaluate this hypothesis, we use the \textbf{CEM} model \citep{sabour2022cem} trained on the EmpatheticDialogues dataset \citep{rashkin2019towards}. \textbf{CEM} conditions on commonsense knowledge generated by \comet{} \citep{bosselut2019comet} to improve empathetic dialogue generation. Using their framework, we apply a \ourdata{}-trained DeBERTa (large) model to filter contextually-relevant fact subsets from the knowledge generated by \comet{}, and use this refined knowledge as the input to the \textbf{CEM} model (denoted as \textbf{CEM w/ \ourdata{}}). 
Our results in Table~\ref{tab:downstream} demonstrate that \textbf{CEM} w/ \ourdata{} outperforms \textbf{CEM} on most metrics, hinting at \ourdata{}'s potential to benefit downstream NLP tasks by enabling improved commonsense knowledge retrieval.\footnote{See Appendix~\ref{apdx:downstream} for more downstream application details.}

\section{Conclusion}
In this work, we propose a general commonsense fact linking task that addresses the challenges of identifying relevant commonsense inferences for textual contexts: contextualization, implicitness, and ambiguity. To promote research into commonsense fact linking, we construct a new, challenging benchmark, \ourdata{}, of over 293k contextually-linked commonsense facts.
Our experimental results show that the predominant heuristic methods used to select relevant commonsense facts for downstream applications perform poorly, motivating the need for new methods that can predict contextually-relevant commonsense inferences.

\section*{Acknowledgements}
We thank Deniz Bayazit, Beatriz Borges, and Mete Ismayilzada for helpful discussions. We also gratefully acknowledge the support of Innosuisse under PFFS-21-29, the EPFL Science Seed Fund, the EPFL Center for Imaging, Sony Group Corporation, and the Allen Institute for AI.

\section*{Limitations}
Specific limitations to this work include the fact that our dataset focuses on short context windows. However, commonsense inferences may connect concepts across longer time windows, which may affect how well models trained on our dataset scale to longer text snippets (as hinted by the lower transfer numbers when models trained on other data subsets are evaluated on \movieatomic{}). 
More broadly, due to the large number of knowledge graphs, narrative corpora, and pretrained models, this work cannot include an exhaustive coverage of this cross-section. Instead, we identify a diverse subset of these resources for our study, though our datasets and methods all use English as a primary language. Finally, some of the models in this work are based on large-scale language models, \eg{}, BERT and RoBERTa, which require considerable resources for inference. In the long run, if neural fact linkers are to be used for large numbers of data snippets (perhaps in online settings), models that can perform faster inference will be needed.

\bibliography{main}
\bibliographystyle{acl_natbib}

\appendix

\begin{table*}[t]
\centering
\resizebox{0.8\textwidth}{!}{
\smallskip\begin{tabular}{lcccc}
\hline
\textbf{Statistic}     &    \personaatomic{}    &    \mutualatomic{}    &      \rocatomic{}     &      \movieatomic{}    \\
\hline
Dialogue/Story Samples &          123           &          237          &          328          &            81          \\
Statements             &          1740          &          1554         &          1640         &           1476         \\
Linked Head Entities   &         17421          &         13088         &         17553         &          20051         \\
Linked Fact Candidates &         72003          &         53120         &         81928         &          86083         \\
\hline
\end{tabular}
}
\caption{Number of sampled contexts and linked fact candidates on the four data portions of \ourdata{}. The context window size is set to 5 where past context $U_{<t}=[U_{t-2},U_{t-1}]$ and future context $U_{>t}=[U_{t+1},U_{t+2}]$.}
\label{tab:collection}
\end{table*}

\begin{table*}[t]
\centering
\resizebox{0.95\textwidth}{!}{
\smallskip\begin{tabular}{lcccccccccccccccc}
\toprule
\multirow{2}*{\textbf{Method}} & \multicolumn{4}{c}{\personaatomic{}} & \multicolumn{4}{c}{\mutualatomic{}} & \multicolumn{4}{c}{\rocatomic{}} & \multicolumn{4}{c}{\movieatomic{}}\\
                        \cmidrule(lr){2-5}  \cmidrule(lr){6-9} \cmidrule(lr){10-13} \cmidrule(lr){14-17}
                        & \textbf{RFC} & \textbf{RHC} & \textbf{IRR} & {$\kappa$} & \textbf{RFC} & \textbf{RHC} & \textbf{IRR} & {$\kappa$} & \textbf{RFC} & \textbf{RHC} & \textbf{IRR} & {$\kappa$} & \textbf{RFC} & \textbf{RHC} & \textbf{IRR} & {$\kappa$} \\
\toprule
\multirow{2}*{Explicit} &  5652  &  1846  &  2963  &\multirow{2}*{0.62} &  2233  &  1692  &  1393  & \multirow{2}*{0.38} & 5172   &  2143  &  2038  & \multirow{2}*{0.48}                         &  6932  &  2853  &  2892  & \multirow{2}*{0.50} \\
                        &  54\%  &  18\%  &  28\%  &                    &  42\%  &  32\%  &  26\%  &                     &  55\%  &  23\%  &  22\%  &                    &  55\%  &  22\%  &  23\%  &                      \\
\midrule
\multirow{2}*{Implicit} &  900  &  1066  &  4994  & \multirow{2}*{0.53} &  1281  &  1697  &  4792  & \multirow{2}*{0.45} &  2530  &  1587  &  4083  & \multirow{2}*{0.59}                         &  1481 &  1326  &  4567  & \multirow{2}*{0.56} \\
                        &  13\%  &  15\%  &  72\%  &                   &  17\%  &  22\%  &  61\%  &                     &  31\%  &  19\%  &  50\%  &                     &  20\%  &  18\%  &  62\%  &                      \\
\midrule
\multirow{2}*{Both}     &  6552  &  2912  &  7957  &\multirow{2}*{0.66} &  3514  &  3389  &  6185  & \multirow{2}*{0.47} &  7702  &  3730  &  6121  & \multirow{2}*{0.57}                         &  8413  &  4179  &  7459  & \multirow{2}*{0.58} \\
                        &  38\%  &  17\%  &  46\%  &                    &  27\%  &  26\%  &  47\%  &                     &  44\%  &  21\%  &  35\%  &                    &  42\%  &  21\%  &  37\%  &                      \\
\bottomrule
\end{tabular}
}
\caption{Relevance validation results of head entities with respect to different fact candidate linking methods. ``RFC'', ``RHC'' and ``IRR'' denote \textit{relevant with full confidence}, \textit{relevant with half confidence} and \textit{irrelevant}, respectively. {$\kappa$} denotes the Cohen's $\kappa$.}
\label{tab:head_stats}
\end{table*}

\begin{table*}[t]
\centering
\resizebox{0.8\textwidth}{!}{
\smallskip\begin{tabular}{llcccccccc}
\hline
\multirow{2}*{\textbf{Type}}     & \multirow{2}*{\textbf{Relation}} & \multicolumn{2}{c}{\personaatomic{}} & \multicolumn{2}{c}{\mutualatomic{}} & \multicolumn{2}{c}{\rocatomic{}} & \multicolumn{2}{c}{\movieatomic{}} \\
                                                    \cmidrule(lr){3-4}      \cmidrule(lr){5-6}      \cmidrule(lr){7-8}      \cmidrule(lr){9-10}
                        &             & \textbf{Count} & \textbf{Percent} & \textbf{Count} & \textbf{Percent} & \textbf{Count} & \textbf{Percent} & \textbf{Count} & \textbf{Percent} \\
\hline
\multirow{8}*{Physical} &  ObjectUse  & 3641  &  32.2\% & 4567  &  40.8\% & 4747  &  28.7\% & 3257  &  26.2\% \\
                        & HasProperty &  834  &   7.4\% &  484  &   4.3\% &  435  &   2.6\% &  535  &   4.3\% \\
                        &  CapableOf  &  424  &   3.7\% &  299  &   2.7\% &  391  &   2.4\% &  614  &   4.9\% \\
                        & AtLocation  &  394  &   3.5\% &  634  &   5.7\% &  375  &   2.3\% &  503  &   4.0\% \\
                        &  MadeUpOf   &  238  &   2.1\% &  336  &   3.0\% &  131  &   0.8\% &  200  &   1.6\% \\
                        &   Desires   &   34  &   0.3\% &  16   &   0.1\% &    8  &  0.05\% &  31   &   0.2\% \\
                        & NotDesires  &   32  &   0.3\% &  16   &   0.1\% &    5  &  0.03\% &  62   &   0.5\% \\
                         \cmidrule(lr){2-10}
                        &    Total    & 5597  &  49.5\% & 6352  &  56.7\% & 6092  &  36.8\% & 5202  &  41.8\% \\
\hline
\multirow{8}*{Event}    & HasSubEvent &  680  &   6.0\% &  530  &   4.7\% &  629  &   3.8\% &  374  &   3.0\% \\
                        & HinderedBy  &  245  &   2.2\% &  184  &   1.6\% &  498  &   3.0\% &  358  &   2.9\% \\
                        &   xReason   &   72  &   0.6\% &  28   &   0.2\% &   35  &   0.2\% &  20   &   0.2\% \\
                        &   Causes    &   29  &   0.3\% &  24   &   0.2\% &   36  &   0.2\% &  75   &   0.6\% \\
                        & isFilledBy  &   26  &   0.2\% &  25   &   0.2\% &   43  &   0.3\% &  29   &   0.2\% \\
                        &  isBefore   &   21  &   0.2\% &  39   &   0.3\% &  109  &   0.7\% &  94   &   0.8\% \\
                        &   isAfter   &   14  &   0.1\% &  27   &   0.2\% &  113  &   0.7\% &  98   &   0.8\% \\
                         \cmidrule(lr){2-10}
                        &    Total    & 1087  &   9.6\% &  857  &   7.6\% & 1463  &   8.8\% & 1048  &   8.4\% \\
\hline
\multirow{10}*{Social}  &    xNeed    & 1470  &  13.0\% & 1306  &  11.7\% & 2883  &  17.4\% & 1280  &  10.3\% \\
                        &    xWant    &  973  &   8.6\% &  766  &   6.8\% & 1982  &  12.0\% & 1143  &   9.2\% \\
                        &   xIntent   &  876  &   7.7\% &  538  &   4.8\% & 1560  &   9.4\% &  764  &   6.1\% \\
                        &   xReact    &  368  &   3.3\% &  253  &   2.3\% &  411  &   2.5\% &  425  &   3.4\% \\
                        &   xEffect   &  361  &   3.2\% &  329  &   2.9\% &  951  &   5.7\% &  912  &   7.3\% \\
                        &    xAttr    &  352  &   3.1\% &  360  &   3.2\% &  369  &   2.2\% &  512  &   4.1\% \\
                        &    oWant    &  111  &   1.0\% &  258  &   2.3\% &  430  &   2.6\% &  440  &   3.5\% \\
                        &   oEffect   &   64  &   0.6\% &  102  &   0.9\% &  287  &   1.7\% &  474  &   3.8\% \\
                        &   oReact    &   56  &   0.5\% &  83   &   0.7\% &  136  &   0.8\% &  252  &   2.0\% \\
                         \cmidrule(lr){2-10}
                        &    Total    & 4631  &  40.9\% & 3995  &  35.7\% & 9009  &  54.4\% & 6202  &  49.8\% \\
\hline
Total                   &             & 11315 & 100.0\% & 11204 & 100.0\% & 16564 & 100.0\% & 12452 & 100.0\% \\
\hline
\end{tabular}
}
\caption{\atomicTT{} relation coverage of relevant facts in \ourdata{}.}
\label{tab:coverage}
\end{table*}

\begin{table*}[t]
\centering
\resizebox{1.0\textwidth}{!}{
\smallskip\begin{tabular}{lcccccccccccccccc}
\hline
\multirow{2}*{\textbf{Split}} & \multicolumn{4}{c}{\personaatomic{}} & \multicolumn{4}{c}{\mutualatomic{}} & \multicolumn{4}{c}{\rocatomic{}} & \multicolumn{4}{c}{\movieatomic{}}\\
                        \cmidrule(lr){2-5}  \cmidrule(lr){6-9} \cmidrule(lr){10-13} \cmidrule(lr){14-17}
      & \# N/D  & \# S/U  &  \# HE  &  \# FC  & \# N/D  & \# S/U  &  \# HE  &  \# FC  & \# N/D  & \# S/U  &  \# HE  &  \# FC  & \# N/D  & \# S/U  &  \# HE  &  \# FC \\
\hline
Train &    90   &   1296  &  12789  &  49526  &   170   &  1147   &  9613   &  33968  &   235   &  1175   &  12520  &  53045  &   58    &  1047   &  14205  &  52690 \\
Valid &    15   &    194  &   1985  &   7912  &    33   &   230   &  1778   &  5896   &    46   &   230   &   2434  &  10403  &   11    &   177   &   2481  &  9307  \\
Test  &    18   &    250  &   2647  &   9758  &    34   &   177   &  1697   &  6328   &    47   &   235   &   2599  &  10854  &   12    &   252   &   3365  &  11646 \\
\hline
Total &   123   &   1740  &  17421  &  67196  &   237   &  1554   &  13088  &  46192  &   328   &  1640   &  17553  &  74302  &   81    &  1476   &  20051  &  73643 \\
\hline
\end{tabular}
}
\caption{Split of narratives or dialogues (N/D) on the four data portions of \ourdata{}, with their contained number of statements or utterances (S/U), labeled head entities (HE) and labeled fact candidates (FC).}
\label{tab:split}
\end{table*}

\begin{table*}[t]
\centering
\resizebox{0.88\textwidth}{!}{
\smallskip\begin{tabular}{cllcccccccc}
\hline
\multirow{2}*{\textbf{Context}} & \multirow{2}*{\textbf{Model}} & \multirow{2}*{\textbf{Setting}} & \multicolumn{4}{c}{\personaatomic{}} & \multicolumn{4}{c}{\rocatomic{}} \\
                                                \cmidrule(lr){4-7}                   \cmidrule(lr){8-11}
                    &                      &                        & \textbf{Accuracy} & \textbf{Precision} & \textbf{Recall} & \textbf{F1} & \textbf{Accuracy} & \textbf{Precision} & \textbf{Recall} & \textbf{F1} \\
\hline
\multirow{20}*{$U_{\leq t}$} & Heuristic (explicit) & \multirow{4}*{none} & 0.451 & 0.247 & 0.779 & 0.375    & 0.501 & 0.258 & 0.616 & 0.364 \\
& Heuristic (implicit)           &              & 0.549 & 0.140 & 0.221 & 0.171        & 0.499 & 0.198 & 0.384 & 0.261 \\
& Heuristic                      &              & 0.211 & 0.211 & \textbf{1.000} & 0.348   & 0.231 & 0.231 & \textbf{1.000} & 0.375 \\
& Head Linking                   &              & 0.600 & 0.334 & 0.899 & 0.487        & 0.537 & 0.326 & 0.936 & 0.484      \\
\cmidrule(lr){2-11}
& \multirow{2}*{LSTM}            & direct       & 0.805 & 0.549 & 0.413 & 0.471        & 0.761 & 0.450 & 0.464 & 0.457      \\
&                                & pipeline     & 0.796 & 0.523 & 0.404 & 0.456        & 0.745 & 0.479 & 0.386 & 0.428      \\
\cmidrule(lr){2-11}
& \multirow{2}*{DistilBERT}      & direct       & 0.840 & 0.617 & 0.636 & 0.626        & 0.811 & 0.584 & 0.635 & 0.608      \\
&                                & pipeline     & 0.834 & 0.615 & 0.568 & 0.591        & 0.791 & 0.540 & 0.657 & 0.593      \\
\cmidrule(lr){2-11}
& \multirow{2}*{BERT (base)}     & direct       & 0.859 & 0.672 & 0.647 & 0.659        & 0.841 & 0.646 & 0.693 & 0.669      \\
&                                & pipeline     & 0.855 & 0.668 & 0.622 & 0.644        & 0.819 & 0.595 & 0.687 & 0.638      \\
\cmidrule(lr){2-11}
& \multirow{2}*{BERT (large)}    & direct       & 0.859 & 0.670 & 0.650 & 0.660        & 0.848 & 0.653 & 0.732 & 0.690      \\
&                                & pipeline     & 0.856 & 0.673 & 0.620 & 0.645        & 0.828 & 0.606 & 0.728 & 0.661      \\
\cmidrule(lr){2-11}
& \multirow{2}*{RoBERTa (base)}  & direct       & 0.866 & 0.691 & 0.657 & 0.674        & 0.845 & 0.645 & 0.734 & 0.687      \\
&                                & pipeline     & 0.859 & 0.663 & 0.679 & 0.671        & 0.831 & 0.609 & 0.748 & 0.671      \\
\cmidrule(lr){2-11}
& \multirow{2}*{RoBERTa (large)} & direct       & 0.883 & 0.735 & 0.698 & 0.716        & 0.874 & 0.709 & 0.774 & 0.740      \\
&                                & pipeline     & 0.874 & 0.706 & 0.690 & 0.698        & 0.861 & 0.673 & 0.776 & 0.721      \\
\cmidrule(lr){2-11}
& \multirow{2}*{DeBERTa (base)}  & direct       & 0.869 & 0.702 & 0.664 & 0.682        & 0.871 & 0.699 & 0.781 & 0.737  \\
&                                & pipeline     & 0.864 & 0.683 & 0.661 & 0.672        & 0.848 & 0.637 & 0.793 & 0.706  \\
\cmidrule(lr){2-11}
& \multirow{2}*{DeBERTa (large)} & direct   & \textbf{0.885} & \textbf{0.743} & 0.693 & \textbf{0.717}  & \textbf{0.884} & \textbf{0.725} & 0.806 & \textbf{0.763}  \\
&                                & pipeline     & 0.873 & 0.691 & 0.722 & 0.706        & 0.868 & 0.678 & 0.816 & 0.741  \\
\cmidrule(lr){1-11}
\multirow{17}*{$C_{t}$} & Head Linking & none  & 0.595 & 0.330 & 0.892 & 0.482      & 0.543 & 0.328 & 0.931 & 0.485      \\
\cmidrule(lr){2-11}
& \multirow{2}*{LSTM}            & direct & 0.818 & 0.615 & 0.369 & 0.461      & 0.766 & 0.491 & 0.365 & 0.419      \\
&                                & pipeline     & 0.813 & 0.605 & 0.350 & 0.443      & 0.756 & 0.470 & 0.436 & 0.452      \\
\cmidrule(lr){2-11}
& \multirow{2}*{DistilBERT}      & direct       & 0.844 & 0.630 & 0.627 & 0.628      & 0.808 & 0.572 & 0.668 & 0.616      \\
&                                & pipeline     & 0.842 & 0.633 & 0.600 & 0.616      & 0.791 & 0.537 & 0.703 & 0.609      \\
\cmidrule(lr){2-11}
& \multirow{2}*{BERT (base)}     & direct       & 0.862 & 0.682 & 0.651 & 0.666      & 0.845 & 0.650 & 0.716 & 0.681      \\
&                                & pipeline     & 0.853 & 0.661 & 0.624 & 0.642      & 0.823 & 0.596 & 0.724 & 0.654      \\
\cmidrule(lr){2-11}
& \multirow{2}*{BERT (large)}    & direct       & 0.869 & 0.704 & 0.657 & 0.680      & 0.855 & 0.667 & 0.743 & 0.703      \\
&                                & pipeline     & 0.854 & 0.666 & 0.622 & 0.643      & 0.833 & 0.610 & 0.767 & 0.680      \\
\cmidrule(lr){2-11}
& \multirow{2}*{RoBERTa (base)}  & direct       & 0.871 & 0.703 & 0.671 & 0.687      & 0.845 & 0.642 & 0.748 & 0.691      \\
&                                & pipeline     & 0.862 & 0.670 & 0.682 & 0.676      & 0.839 & 0.627 & 0.750 & 0.683      \\
\cmidrule(lr){2-11}
& \multirow{2}*{RoBERTa (large)} & direct   & 0.882 & 0.721 & 0.720 & \textbf{0.721} & 0.879 & \textbf{0.720} & 0.779 & 0.748 \\
&                                & pipeline     & 0.874 & 0.716 & 0.691 & 0.703      & 0.867 & 0.687 & 0.781 & 0.731       \\
\cmidrule(lr){2-11}
& \multirow{2}*{DeBERTa (base)}  & direct       & 0.871 & 0.701 & 0.676 & 0.688        & 0.859 & 0.661 & 0.799 & 0.723  \\
&                                & pipeline     & 0.864 & 0.683 & 0.660 & 0.671        & 0.853 & 0.644 & 0.809 & 0.718  \\
\cmidrule(lr){2-11}
& \multirow{2}*{DeBERTa (large)} & direct  & \textbf{0.884} & \textbf{0.734} & 0.707 & 0.720    & \textbf{0.884} & 0.717 & 0.826 & \textbf{0.768}  \\
&                                & pipeline     & 0.880 & 0.730 & 0.684 & 0.707        & 0.880 & 0.712 & 0.806 & 0.756  \\
\hline
\end{tabular}
}
\caption{Fact linking results on \personaatomic{} and \rocatomic{}.}
\label{tab:full_main_1}
\end{table*}

\begin{table*}[t]
\centering
\resizebox{0.88\textwidth}{!}{
\smallskip\begin{tabular}{cllcccccccc}
\hline
\multirow{2}*{\textbf{Context}} & \multirow{2}*{\textbf{Model}} & \multirow{2}*{\textbf{Setting}} & \multicolumn{4}{c}{\mutualatomic{}} & \multicolumn{4}{c}{\movieatomic{}} \\
                                                \cmidrule(lr){4-7}                   \cmidrule(lr){8-11}
                    &                      &                        & \textbf{Accuracy} & \textbf{Precision} & \textbf{Recall} & \textbf{F1} & \textbf{Accuracy} & \textbf{Precision} & \textbf{Recall} & \textbf{F1} \\
\hline
\multirow{20}*{$U_{\leq t}$} & Heuristic (explicit) & \multirow{4}*{none} & 0.555 & 0.344 & 0.583 & 0.433    & 0.487 & 0.247 & 0.738 & 0.370 \\
& Heuristic (implicit)           &              & 0.445 & 0.239 & 0.417 & 0.304        & 0.513 & 0.138 & 0.262 & 0.181      \\
& Heuristic                      &          & 0.290 & 0.290 & \textbf{1.000} & 0.450   & 0.205 & 0.205 & \textbf{1.000} & 0.340      \\
& Head Linking                   &              & 0.643 & 0.445 & 0.931 & 0.602        & 0.548 & 0.301 & 0.912 & 0.452      \\
\cmidrule(lr){2-11}
& \multirow{2}*{LSTM}            & direct       & 0.749 & 0.566 & 0.581 & 0.573        & 0.769 & 0.431 & 0.404 & 0.417      \\
&                                & pipeline     & 0.745 & 0.557 & 0.596 & 0.576        & 0.761 & 0.409 & 0.374 & 0.391      \\
\cmidrule(lr){2-11}
& \multirow{2}*{DistilBERT}      & direct       & 0.792 & 0.636 & 0.663 & 0.649        & 0.800 & 0.510 & 0.514 & 0.512      \\
&                                & pipeline     & 0.773 & 0.602 & 0.651 & 0.625        & 0.796 & 0.501 & 0.509 & 0.505      \\
\cmidrule(lr){2-11}
& \multirow{2}*{BERT (base)}     & direct       & 0.801 & 0.648 & 0.690 & 0.668        & 0.818 & 0.558 & 0.537 & 0.547      \\
&                                & pipeline     & 0.781 & 0.616 & 0.656 & 0.635        & 0.814 & 0.547 & 0.534 & 0.540      \\
\cmidrule(lr){2-11}
& \multirow{2}*{BERT (large)}    & direct       & 0.819 & 0.672 & 0.706 & 0.689        & 0.835 & 0.598 & 0.593 & 0.595      \\
&                                & pipeline     & 0.797 & 0.634 & 0.711 & 0.670        & 0.827 & 0.575 & 0.588 & 0.582      \\
\cmidrule(lr){2-11}
& \multirow{2}*{RoBERTa (base)}  & direct       & 0.810 & 0.663 & 0.703 & 0.682        & 0.820 & 0.561 & 0.546 & 0.553      \\
&                                & pipeline     & 0.794 & 0.631 & 0.700 & 0.664        & 0.818 & 0.551 & 0.543 & 0.547      \\
\cmidrule(lr){2-11}
& \multirow{2}*{RoBERTa (large)} & direct       & 0.835 & 0.702 & 0.748 & 0.724    & \textbf{0.850} & \textbf{0.635} & 0.628 & 0.631   \\
&                                & pipeline     & 0.819 & 0.658 & 0.787 & 0.717        & 0.834 & 0.590 & 0.623 & 0.606      \\
\cmidrule(lr){2-11}
& \multirow{2}*{DeBERTa (base)}  & direct       & 0.836 & 0.694 & 0.779 & 0.734        & 0.820 & 0.562 & 0.551 & 0.556  \\
&                                & pipeline     & 0.822 & 0.665 & 0.781 & 0.719        & 0.817 & 0.553 & 0.549 & 0.551  \\
\cmidrule(lr){2-11}
& \multirow{2}*{DeBERTa (large)} & direct   & \textbf{0.861} & \textbf{0.751} & 0.781 & \textbf{0.766}   & \textbf{0.850} & 0.621 & 0.683 & \textbf{0.651}  \\
&                                & pipeline     & 0.835 & 0.671 & 0.847 & 0.749        & 0.839 & 0.596 & 0.656 & 0.625  \\
\cmidrule(lr){1-11}
\multirow{17}*{$C_{t}$} & Head Linking & none  & 0.647 & 0.447 & 0.913 & 0.600        & 0.564 & 0.307 & 0.901 & 0.458      \\
\cmidrule(lr){2-11}
& \multirow{2}*{LSTM}            & direct       & 0.755 & 0.577 & 0.585 & 0.581        & 0.772 & 0.437 & 0.393 & 0.414      \\
&                                & pipeline     & 0.756 & 0.581 & 0.579 & 0.580        & 0.771 & 0.434 & 0.388 & 0.409      \\
\cmidrule(lr){2-11}
& \multirow{2}*{DistilBERT}      & direct       & 0.795 & 0.641 & 0.667 & 0.654        & 0.807 & 0.527 & 0.536 & 0.531      \\
&                                & pipeline     & 0.781 & 0.620 & 0.638 & 0.629        & 0.801 & 0.514 & 0.513 & 0.514      \\
\cmidrule(lr){2-11}
& \multirow{2}*{BERT (base)}     & direct       & 0.804 & 0.653 & 0.694 & 0.673        & 0.829 & 0.587 & 0.562 & 0.574      \\
&                                & pipeline     & 0.795 & 0.638 & 0.682 & 0.659        & 0.814 & 0.544 & 0.555 & 0.550      \\
\cmidrule(lr){2-11}
& \multirow{2}*{BERT (large)}    & direct       & 0.817 & 0.670 & 0.727 & 0.697        & 0.839 & 0.607 & 0.604 & 0.605      \\
&                                & pipeline     & 0.797 & 0.639 & 0.688 & 0.663        & 0.824 & 0.567 & 0.597 & 0.582      \\
\cmidrule(lr){2-11}
& \multirow{2}*{RoBERTa (base)}  & direct       & 0.811 & 0.662 & 0.716 & 0.688        & 0.832 & 0.590 & 0.588 & 0.589      \\
&                                & pipeline     & 0.795 & 0.638 & 0.682 & 0.659        & 0.809 & 0.530 & 0.576 & 0.552      \\
\cmidrule(lr){2-11}
& \multirow{2}*{RoBERTa (large)} & direct       & 0.838 & 0.694 & 0.792 & 0.740     & \textbf{0.851} & \textbf{0.637} & 0.633 & 0.635      \\
&                                & pipeline     & 0.825 & 0.671 & 0.780 & 0.722        & 0.830 & 0.578 & 0.631 & 0.603      \\
\cmidrule(lr){2-11}
& \multirow{2}*{DeBERTa (base)}  & direct       & 0.842 & 0.712 & 0.766 & 0.738        & 0.842 & 0.618 & 0.601 & 0.609  \\
&                                & pipeline     & 0.824 & 0.661 & 0.805 & 0.726        & 0.805 & 0.520 & 0.619 & 0.565  \\
\cmidrule(lr){2-11}
& \multirow{2}*{DeBERTa (large)} & direct   & \textbf{0.859} & \textbf{0.726} & 0.826 & \textbf{0.773}   & 0.848 & 0.620 & 0.657 & \textbf{0.638}  \\
&                                & pipeline     & 0.848 & 0.705 & 0.817 & 0.757        & 0.839 & 0.595 & 0.666 & 0.628  \\
\hline
\end{tabular}
}
\caption{Fact linking results on \mutualatomic{} and \movieatomic{}.}
\label{tab:full_main_2}
\end{table*}

\begin{table*}[t]
\centering
\resizebox{0.8\textwidth}{!}{
\smallskip\begin{tabular}{clcccccccc}
\hline
\multirow{2}*{\textbf{Context}} & \multirow{2}*{\textbf{Model}} & \multicolumn{4}{c}{\personaatomic{}} & \multicolumn{4}{c}{\rocatomic{}} \\
                                             \cmidrule(lr){3-6}                   \cmidrule(lr){7-10}
        &          & \textbf{Accuracy} & \textbf{Precision} & \textbf{Recall} & \textbf{F1} & \textbf{Accuracy} & \textbf{Precision} & \textbf{Recall} & \textbf{F1} \\
\hline
\multirow{6}*{$U_{\leq t}$}  & LSTM            & 0.683 & 0.767 & 0.612 & 0.681 & 0.704 & 0.752 & 0.795 & 0.773 \\
                  & DistilBERT      & 0.787 & 0.779 & 0.856 & 0.816 & 0.764 & 0.780 & 0.875 & 0.825 \\
                  & BERT (base)     & 0.802 & 0.803 & 0.852 & 0.827 & 0.766 & 0.782 & 0.874 & 0.825 \\
                  & BERT (large)    & 0.794 & 0.796 & 0.845 & 0.820 & 0.772 & 0.782 & 0.888 & 0.832 \\
                  & RoBERTa (base)  & 0.823 & 0.830 & 0.854 & 0.842 & 0.783 & 0.802 & 0.871 & 0.835 \\
                  & RoBERTa (large) & \textbf{0.834} & 0.834 & 0.874 & \textbf{0.854} & 0.819 & 0.830 & \textbf{0.898} & \textbf{0.863} \\
\hline
\multirow{6}*{$C_{t}$}  & LSTM            & 0.688 & 0.759 & 0.638 & 0.693 & 0.703 & 0.744 & 0.809 & 0.775 \\
                  & DistilBERT      & 0.794 & 0.783 & 0.867 & 0.823 & 0.758 & 0.771 & 0.879 & 0.821 \\
                  & BERT (base)     & 0.793 & 0.791 & 0.852 & 0.820 & 0.773 & 0.787 & 0.879 & 0.830 \\
                  & BERT (large)    & 0.797 & 0.804 & 0.837 & 0.820 & 0.780 & 0.792 & 0.885 & 0.836 \\
                  & RoBERTa (base)  & 0.825 & \textbf{0.850} & 0.829 & 0.839 & 0.791 & 0.817 & 0.863 & 0.839 \\
                  & RoBERTa (large) & 0.832 & 0.830 & \textbf{0.875} & 0.852 & \textbf{0.821} & \textbf{0.836} & 0.892 & \textbf{0.863} \\
\hline
\end{tabular}
}
\caption{Head entity linking results on \personaatomic{} and \rocatomic{}, given different context windows $U_{\leq t}=[U_{<t},U_{t}]$ or $C_{t}=[U_{<t},U_{t},U_{>t}]$.}
\label{tab:pipe1}
\end{table*}

\begin{table*}[t]
\centering
\resizebox{0.8\textwidth}{!}{
\smallskip\begin{tabular}{clcccccccc}
\hline
\multirow{2}*{\textbf{Context}} & \multirow{2}*{\textbf{Model}} & \multicolumn{4}{c}{\personaatomic{}} & \multicolumn{4}{c}{\rocatomic{}} \\
                                             \cmidrule(lr){3-6}                   \cmidrule(lr){7-10}
        &          & \textbf{Accuracy} & \textbf{Precision} & \textbf{Recall} & \textbf{F1} & \textbf{Accuracy} & \textbf{Precision} & \textbf{Recall} & \textbf{F1} \\
\hline
\multirow{6}*{$U_{\leq t}$} & LSTM            & 0.710 & 0.719 & 0.470 & 0.568 & 0.694 & 0.622 & 0.554 & 0.586 \\
                  & DistilBERT      & 0.789 & 0.779 & 0.671 & 0.721 & 0.774 & 0.703 & 0.730 & 0.716 \\
                  & BERT (base)     & 0.813 & 0.810 & 0.705 & 0.754 & 0.811 & 0.746 & 0.783 & 0.764 \\
                  & BERT (large)    & 0.815 & 0.811 & 0.706 & 0.755 & 0.824 & 0.766 & 0.795 & 0.780 \\
                  & RoBERTa (base)  & 0.836 & 0.803 & 0.789 & 0.796 & 0.823 & 0.754 & 0.812 & 0.782 \\
                  & RoBERTa (large) & 0.846 & 0.845 & 0.760 & 0.800 & 0.853 & 0.803 & 0.827 & 0.815 \\
\hline
\multirow{6}*{$C_{t}$} & LSTM            & 0.718 & 0.753 & 0.456 & 0.568 & 0.695 & 0.636 & 0.514 & 0.569 \\
                  & DistilBERT      & 0.797 & 0.792 & 0.678 & 0.731 & 0.782 & 0.702 & 0.769 & 0.734 \\
                  & BERT (base)     & 0.813 & 0.811 & 0.705 & 0.754 & 0.823 & 0.757 & 0.804 & 0.780 \\
                  & BERT (large)    & 0.819 & 0.821 & 0.707 & 0.760 & 0.832 & 0.764 & 0.826 & 0.794 \\
                  & RoBERTa (base)  & 0.833 & 0.794 & \textbf{0.796} & 0.795 & 0.832 & 0.763 & 0.829 & 0.795 \\
                  & RoBERTa (large) & \textbf{0.850} & \textbf{0.846} & 0.770 & \textbf{0.806} & \textbf{0.864} & \textbf{0.819} & \textbf{0.838} & \textbf{0.828} \\
\hline
\end{tabular}
}
\caption{Fact linking results of head entities classified as relevant on \personaatomic{} and \rocatomic{}, given given different context windows $U_{\leq t}=[U_{<t},U_{t}]$ or $C_{t}=[U_{<t},U_{t},U_{>t}]$.}
\label{tab:pipe2}
\end{table*}

\begin{table}[t]
\centering
\resizebox{\columnwidth}{!}{
\smallskip\begin{tabular}{lrrrr}
\toprule
\textbf{Fact Type} & \textbf{Acc.} & \textbf{Prec.} & \textbf{Recall} & \textbf{F1} \\
\toprule
physical      & 0.888 & 0.775 & 0.753 & 0.764 \\
event\&social & 0.879 & 0.691 & 0.639 & 0.664 \\
\midrule
all           & 0.883 & 0.735 & 0.698 & 0.716 \\
\bottomrule
\end{tabular}
}
\caption{Fact linking results of \textbf{RoBERTa (large)} on \personaatomic{} with respect to different fact types, under the direct setting and window $U_{\leq t}=[U_{<t},U_{t}]$.}
\label{tab:type}
\end{table}

\section{Data Collection Details}
\label{apdx:collection}

\paragraph{Dataset Selection} The narrative contexts in \ourdata{} are sampled from four stylistically diverse dialogue and storytelling datasets.
Specifically, \personachat{} \cite{zhang2018personalizing} contains a rich amount of consistent chit-chat dialogues crowdsourced with additional persona profiles.
MuTual \citep{cui2020mutual} contains more reasoning-focused dialogues from English listening comprehension exams.
ROCStories \cite{mostafazadeh2016corpus} and the CMU Movie Summary Corpus \citep{bamman2013learning} are commonly used storytelling datasets.
All above datasets involve elaborate contextual inference and understanding.

For \personachat{}, MuTual and ROCStories, we sample dialogues or stories which (potentially) involve the richest commonsense knowledge.
In particular, we conduct fact candidate linking (as described in Sec.~\ref{sec:hfcl}) on all dialogues and stories in these datasets, and then select the dialogues and stories that have the most fact candidates.
For the CMU Movie Summary Corpus, we sample movie summaries that belong to the genre of \textit{slice of life story}, \textit{childhood drama}, \textit{children's} and/or \textit{family}, which are supposed to involve more commonsense inferences, and meanwhile remove movie summaries that also belong to non-commonsensical genres, \eg{}, \textit{fantasy}, \textit{supernatural}, \textit{mystery}, etc.
The total number of our sampled dialogues/stories, statements, and linked head entities and fact candidates are summarized in Table~\ref{tab:collection}.

\paragraph{Knowledge Graph} We use \atomicTT{} \citep{hwang2021comet} as the commonsense knowledge graph for building \ourdata{}.
This advanced knowledge graph contains 1.33M everyday inferential facts covering a rich variety of complex entities, where 0.21M facts are about physical objects, 0.20M facts are centered on daily events, and other 0.92M facts involve social interactions.

\paragraph{Crowdsourcing Details} We also conduct worker qualifications for our crowdsourcing relevance judgements described in Sec.~\ref{sec:cvv}.
Specifically, for head entity validation, we test workers with 5 narrative contexts, each with 4 linked head entities, and choose workers who can annotate 19 or more (\ie{}, $\geq 95\%$) head entities reasonably.
For fact candidate validation, we still test workers with 5 narrative contexts, each with 4 linked fact candidates, and choose workers who can annotate 18 or more (\ie{}, $\geq 90\%$) fact candidates reasonably.
The number of workers that we choose as qualified for head entity and fact candidate validation are 54 and 106, respectively.
For \personaatomic{}, \mutualatomic{} and \rocatomic{}, we pay each worker \$1.20 for every 60 annotations in the head entity validation, and \$1.00, \$1.60 and \$2.00 for every 60 annotations in the three rounds of fact candidate validation, respectively.
For \movieatomic{}, which involves more complex narratives, we pay each worker \$1.80 for every 60 annotations in the head entity validation, and \$1.50, \$2.50 and \$3.50 for every 60 annotations in the three rounds of fact candidate validation, respectively.
The average hourly wage for each worker is about \$25.00.

\section{Data Statistics Details}
\label{apdx:statistics}
Table~\ref{tab:head_stats} shows stratified statistics of the crowdsourced head entity relevance annotations for the different candidate linking methods described in Sec.~\ref{sec:hfcl} (\ie{}, explicit pattern matching, implicit embedding matching).
For contextual relevance, we draw similar conclusions as the statistics of fact relevance annotations described in Sec.~\ref{data_analysis}.
While in terms of implicitness, different from the statistics of fact relevance annotations, we find that implicit head entities retrieved using embedding similarity are less likely to be relevant to the context than pattern-matched head entities.
This shows that implicitly related head entities are more difficult to be retrieved than the explicitly related head entities.
We also observe that relevance validation on head entities has overall lower Cohen's $\kappa$ than that of fact candidates shown in Table~\ref{tab:fact_stats}.
This indicates that linked head entities contain more relevance controversy compared to their fact candidates, since a head node is more vague than the whole fact it relates to, which provides less information that increases the ambiguity in relevance.
Besides, the Cohen's $\kappa$ ranking of explicit and implicit (and both) methods in head entity linking seems to change randomly across different data portions in \ourdata{}.
This implies that the difference of ambiguity between explicit and implicit linking narrows down as the linked object becomes simpler (\ie{}, from a whole fact triple to a head node in it).

We also investigate the coverage of \textit{always} and \textit{sometimes relevant} facts in \ourdata{} on the relations of \atomicTT{} knowledge graph.
The statistical results are shown in Table~\ref{tab:coverage}, where different \atomicTT{} relations are associated with different commonsense knowledge types.
Besides the simplest knowledge of ObjectUse, we find that social interaction of ``PersonX'' (xNeed, xWant, xIntent, xReact, xEffect and xAttr) occupy a large proportion of relevant facts.
And in general, more than half of the relevant facts are beyond simple physical knowledge, which involve more complicated daily events or social interactions.
This shows that context inference and understanding widely involve complicated daily events and social knowledge, which are difficult to be retrieved by simple heuristics.
This reveals the necessity of improving commonsense fact linking methods in NLP systems.

\section{Experimental Details}
\label{apdx:exp}
Table~\ref{tab:split} shows our split of training, development and testing sets on the four data portions of \ourdata{}. Note that the total number of labeled fact candidates does not match up with Table~\ref{tab:fact_stats} because we remove at odds facts and include facts of irrelevant head entities which are not validated in crowdsourcing.

For the fact linker based on two-layer bidirectional LSTM, we use GloVe \cite{pennington2014glove} to initialize the embedding matrix, and set embedding size, hidden size and vocabulary size as 300, 300 and 10000, respectively.
The LSTM encoder is combined with an MLP classifier on top of the output hidden state concatenation of the start token ``<bos>'' and the end token ``<eos>'', where the MLP inner layer size is set as 1200.
We set dropout rate as 0.5 and use Adam optimizer \cite{kingma2014adam} with a learning rate of $2e^{-4}$, which is selected via grid search from $\{5e^{-5},1e^{-4},2e^{-4},5e^{-4},1e^{-3}\}$.
For the fact linkers based on pretrained language models, we use the default model settings on Hugging Face\footnote{\url{https://huggingface.co}}.
All pretrained language models are combined with a linear classifier on top of the output hidden states of the start token (``[CLS]'' for BERT and DistilBERT, ``<s>'' for RoBERTa).
Adam optimizer is still used and the learning rates are set as $2e^{-6}$ for DistilBERT, BERT (base) and RoBERTa (base), and $5e^{-7}$ for BERT (large) and RoBERTa (large), selecting via grid search from $\{1e^{-7},2e^{-7},5e^{-7},1e^{-6},2e^{-6},5e^{-6},1e^{-5}\}$.
All models are trained using binary cross entropy loss.

On each data portion of \ourdata{}, we train each fact linker for 20 epochs and test the performance of the fact linker from the epoch where it achieves the best F1 score on the validation set.
The training and evaluation batch sizes are both set as 8 for LSTM, DistilBERT, BERT (base) and RoBERTa (base), and 2 for BERT (large) and RoBERTa (large).
Model training and evaluation is performed on four NVIDIA TITAN X Pascal GPUs.

\paragraph{Recall Measurement} Finally, we note that our F1 scores depend on a faithful measurement of recall. This measurement poses a recurring challenge in open-ended retrieval tasks as measuring recall requires approximating of the ``true'' number of relevant facts to a narrative context using a suitable gold set of facts. In our case, the full \atomicTT{} KG would be the most expansive candidate set, but using it as a gold set would require annotating 1M$+$ facts for each narrative context, which is not scalable. Instead, we record recall with respect to a concrete set of candidates that heuristics initially over-sample, $\sim$41 facts per example context (excluding the facts which are annotated as \textit{at odds} by crowdworkers), producing a diverse set of initial facts with which to measure recall (\ie, models would not score highly on F1 simply by making few predictions). 
Conceptually, we also note that commonsense KGs only provide a very limited snapshot of commonsense in the world. Therefore, even the full \atomicTT{} KG would not provide a complete picture of relevant commonsense knowledge for measuring recall.

\section{Full Results of Fact Linking}
\label{apdx:full}
Table~\ref{tab:full_main_1} and \ref{tab:full_main_2} shows the full evaluation results of our fact linking baselines on \ourdata{}.
In terms of the comparisons between different models, between direct and pipeline settings, and between context windows $U_{\leq t}$ and $C_{t}$, we draw similar conclusions as described in Sec.~\ref{results}.

\section{Results of Fact Linking Sub-Tasks in Pipeline Setting}
\label{apdx:pipe}
Table~\ref{tab:pipe1} and \ref{tab:pipe2} show the evaluation results of the fact linking sub-tasks in the pipeline prediction setting, including head entity linking and fact linking of head entities classified as relevant.
Experiments are conducted on the \personaatomic{} and \rocatomic{} data portions of \ourdata{}.
For both sub-tasks, we find that all language models achieve overall higher evaluation results compared to their results in the direct prediction setting shown in Table~\ref{tab:main}.
This shows that the pipeline prediction successfully divides the whole fact linking task into two simpler steps.
However, as described in Sec.~\ref{results}, this does not make the pipeline prediction finally outperforms the direct prediction, as error propagation exits between the two sub-tasks.

\section{Knowledge Graph Generalization}
\label{apdx:knowledge_type}
Table~\ref{tab:type} shows the fine-grained fact linking results of RoBERTa (large) with respect to different \atomicTT{} fact types.
In particular, we evaluate RoBERTa (large) model (finetuned on universal training set) on different test subsets including physical facts, event-based and social (event\&social) facts, or all facts (\ie{}, universal test set).
The promising performance of RoBERTa (large) on linking physical knowledge reveals that fact linkers developed on \ourdata{} has the potential of generalizing to other knowledge graphs, \eg{}, ConceptNet whose physical facts make up part of the \atomicTT{} contents.
We also observe that evaluation scores of linking event\&social facts are overall lower than linking physical facts.
This indicates that more elaborate research is needed to make our trained fact linkers generalize to event-based and social knowledge graphs, since they typically involve more complex contents and are often linked to the context in more implicit ways.

\section{Data Examples}
\label{apdx:example}
Table~\ref{tab:examples} shows examples from the four data portions of \ourdata{}, including a piece of context from each data portion and its linked facts with different link types.
As shown in examples, \movieatomic{} has more complex narrative contexts, which contains longer statements compared to the other three data portions.

\begin{table*}[t]
\centering
\resizebox{1.0\textwidth}{!}{
\smallskip\begin{tabular}{cl}
\hline
\multicolumn{2}{c}{\textbf{\personaatomic{}}} \\
\hline
\multirow{5}*{Context} & $U_{t-2}$: I like cooking macrobiotic and healthy food and working out at the gym. \\
                       & $U_{t-1}$: What is macrobiotic food? My best friend is my mother. \\
                       & $U_{t}$\quad: Things like whole grains. I drink at bars so I have to stay healthy. \\
                       & $U_{t+1}$: You should not drink a lot, it's bad for you. \\
                       & $U_{t+2}$: Well that is where I meet women, at bars. So I end up drinking. \\
\cmidrule(lr){1-2}
\multirow{4}*{Fact}    & \textbf{RPA}: stay healthy, HasSubEvent, eat healthy foods (\textit{always relevant}) \\
                       & \textbf{RPP}: stay healthy, xNeed, exercise and eat balanced meals (\textit{always relevant}) \\
                       & \textbf{RPF}: bar, ObjectUse, take their friends to (\textit{sometimes relevant}) \\
                       & \textbf{IRR}\,: PersonX likes to drink, xAttr, thirsty (\textit{irrelevant}) \\
\hline
\multicolumn{2}{c}{\textbf{\mutualatomic{}}} \\
\hline
\multirow{5}*{Context} & $U_{t-2}$: It's \$800 in all, sir. Do you want to pay in cash? \\
                       & $U_{t-1}$: Well, can I use my check please? \\
                       & $U_{t}$\quad: Sorry, sir. We don't take checks. You can pay by credit card. \\
                       & $U_{t+1}$: OK. Here's my credit card.\\
                       & $U_{t+2}$: Thank you, sir. Here you go with your credit card and the receipt.\\
\cmidrule(lr){1-2}
\multirow{4}*{Fact}    & \textbf{RPA}: credit, ObjectUse, pay for the food (\textit{sometimes relevant}) \\
                       & \textbf{RPP}: pay by check, HasSubEvent, know amount of check (\textit{always relevant})\\
                       & \textbf{RPF}: card, ObjectUse, give to clerk (\textit{always relevant}) \\
                       & \textbf{IRR}\,: personal check, ObjectUse, pay someone back (\textit{irrelevant}) \\
\hline
\multicolumn{2}{c}{\textbf{\rocatomic{}}} \\
\hline
\multirow{5}*{Context} & $U_{t-2}$: Jamie was sleeping at a friend's house.  \\
                       &  $U_{t-1}$: It was her first time away at a friend's house. \\
                       &  $U_{t}$\quad: Jamie was scared and missed her home and family. \\
                       &  $U_{t+1}$: She called her mom to pick her up. \\
                       &  $U_{t+2}$: Jamie went home to sleep in her own bed. \\
\cmidrule(lr){1-2}
\multirow{4}*{Fact}    & \textbf{RPA}: PersonX misses PersonX's parents, xNeed, to be away from the parents (\textit{always relevant}) \\
                       & \textbf{RPP}: family, AtLocation, house (\textit{always relevant}) \\
                       & \textbf{RPF}: PersonX feels homesick, xEffect, take leave to go home (\textit{always relevant}) \\
                       & \textbf{IRR}\,: PersonX misses home, xWant, to watch a movie about home (\textit{irrelevant}) \\
\hline
\multicolumn{2}{c}{\textbf{\movieatomic{}}} \\
\hline
\multirow{8}*{Context} & $U_{t-2}$: Fred is surprised at how real Wilma's tears are during a major scene,\\
                       &\quad\quad\; and soon learns she refuses to speak to him for forgetting pebbles.\\
                       & $U_{t-1}$: Fred soon starts living the play and realizes that Christmas isn't about greed, but about happiness and love.\\
                       & $U_{t}$\quad: He soon apologizes to everyone he was rude to, including Wilma, who is still mad at him\\
                       &\quad\quad\; for his thoughtlessness, and everyone in bedrock truly has a very merry Christmas.\\
                       & $U_{t+1}$: But then Fred gets sick however because of the bedrock bug.\\
                       & $U_{t+2}$: This is further complicated when Wilma assures him that he will recover just in time\\
                       &\quad\quad\; to attend his mother-in-law's Christmas dinner.\\
\cmidrule(lr){1-2}
\multirow{4}*{Fact}    & \textbf{RPA}: apologize, xIntent, made mistake (\textit{always relevant}) \\
                       & \textbf{RPP}: PersonX rude to PersonY, oEffect, starts crying (\textit{always relevant}) \\
                       & \textbf{RPF}: PersonX tells Fred, xIntent, Fred to know (\textit{always relevant}) \\
                       & \textbf{IRR}\,: Christmas, HasProperty, celebrated by Christians (\textit{irrelevant}) \\
\hline
\end{tabular}
}
\caption{Data examples in \ourdata{}.}
\label{tab:examples}
\end{table*}

\section{Downstream Application Details}
\label{apdx:downstream}
We use the CEM \citep{sabour2022cem} model trained on the EmpatheticDialogues \citep{rashkin2019towards} dataset as our downstream framework.
EmpatheticDialogues is a large-scale multi-turn dialogue dataset containing 25K empathetic conversations between crowdworkers.
The task of dialogue models on this dataset is to play the role of a listener and generate empathetic responses to a speaker. To augment empathetic response generation with commonsense knowledge, the CEM model first appends five kinds of \atomic{} \citep{sap2019atomic} relation tokens (\textit{xReact}, \textit{xIntent}, \textit{xNeed}, \textit{xEffect} and \textit{xWant}) to each dialogue context.
Based on the context and appended tokens, it then uses \comet{} \citep{bosselut2019comet} to generate five commonsense inferences (\ie{}, tail entities) for each relation, whose purpose is to help the model generate more empathetic responses.

However, the commonsense inferences generated by \comet{} may not all be relevant to the dialogue context and helpful for generating more empathetic responses. 
Therefore, we train neural fact linkers on \ourdata{} to refine the knowledge generated by \comet{}.
Specifically, we train a DeBERTa (large) fact linker on the union of all four data portions of \ourdata{}.
To adapt our fact linker to the CEM setting, where tail entities are generated without given head entities, we remove the head entity of each fact out of the input when training the fact linker.
We also only use the facts whose relations are one of the five CEM appended relations to build our training samples.
Finally, we follow the same hyper-parameter settings suggested by CEM to train the dialogue model (\ie{}, we do not re-tune the model for our setting), and filter out the \comet{} inferences which are classified as irrelevant by our trained fact linker, resulting in $\sim$38.5\% generated facts being removed from the CEM input. For evaluation, we use the same metrics Perplexity (PPL) and Distinct-n \citep{li2016diversity} from \citet{sabour2022cem}, and also include commonly used metrics BLEU \citep{papineni2002bleu}, METEOR \citep{banerjee2005meteor}, ROUGE \citep{lin2004rouge}, CIDEr \citep{vedantam2015cider} and SkipThoughts \citep{kiros2015skip}. Table~\ref{tab:downstream_examples} shows examples of our downstream dialogue response generation results.

\begin{table*}[t]
\centering
\resizebox{0.95\textwidth}{!}{
\smallskip\begin{tabular}{ll}
\hline
\multirow{1}*{Context}              & I could not wait to go to the concert. \\
\hline
\multirow{1}*{CEM}                  & What did you do? \\
\multirow{1}*{CEM w/ \ourdata{}}    & What band is it?  \\
\multirow{1}*{Human}                & Which concert? \\
\hline
\hline
\multirow{1}*{Context}              & My landlady was really understanding when I told her that I would be late with rent this month. \\
\hline
\multirow{1}*{CEM}                  & That is good, did you find out who did it? \\
\multirow{1}*{CEM w/ \ourdata{}}    & That is so nice of her.  \\
\multirow{1}*{Human}                & That is nice. I could get that here. \\
\hline
\hline
\multirow{1}*{Context}              & Running my first (and maybe only!) marathon felt like such a huge accomplishment! \\
\hline
\multirow{1}*{CEM}                  & I am sure you will do great! \\
\multirow{1}*{CEM w/ \ourdata{}}    & That is great! Congratulations! \\
\multirow{1}*{Human}                & Wow, that is an amazing accomplishment! Congratulations! \\
\hline
\end{tabular}
}
\caption{Examples of downstream dialogue response generation results on the EmpatheticDialogues dataset.}
\label{tab:downstream_examples}
\end{table*}

\end{document}